\PassOptionsToPackage{table}{xcolor} 
\documentclass[]{fairmeta}

\usepackage{silence}
\usepackage[utf8]{inputenc} 
\usepackage[T1]{fontenc}    
\usepackage{url}            
\usepackage{booktabs}       
\usepackage{amsfonts}       
\usepackage{nicefrac}       
\usepackage{microtype}      
\usepackage{xcolor}
\usepackage{soul}
\definecolor{bluelink}{RGB}{0,113,188}
\definecolor{greenlink}{RGB}{0,188,113}
\definecolor{PineGreen}{RGB}{0.0, 0.47, 0.44}
\definecolor{Gray}{RGB}{0.5,0.5,0.5}
\definecolor{audio_desc}{rgb}{0.82,0.99,0.80}
\definecolor{shot_desc_1}{rgb}{0.98,0.87,0.87}
\definecolor{shot_desc_2}{rgb}{0.98,0.95,0.83}
\definecolor{shot_desc_3}{rgb}{0.92,0.86,0.98}
\usepackage{listings} 
\usepackage{wrapfig}
\usepackage[most]{tcolorbox}

\newtcolorbox{samplebox}[1]{
    breakable, 
    colback=blue!5!white,
    colframe=blue!50!black, 
    fonttitle=\bfseries,
    title=#1
}
\definecolor{citecolor}{HTML}{0071bc}
\hypersetup{
    colorlinks=true,%
    citecolor=citecolor,%
    filecolor=citecolor,%
    linkcolor=citecolor,%
    urlcolor=citecolor
}
\usepackage{tabularx}
\usepackage[most]{tcolorbox}
\usepackage{amsmath}
\usepackage{multirow}
\usepackage{array}
\usepackage{wrapfig}
\usepackage{enumitem}
\usepackage{tikz}
\usepackage{lipsum}
\usepackage{adjustbox}
\usepackage{siunitx}          
\usepackage{graphicx}     
\usepackage{float}

\usepackage{amssymb}
\renewcommand{\paragraph}[1]{\vspace{1.25mm}\noindent\textbf{#1}}

\newlength\savewidth
\newcolumntype{x}[1]{>{\centering\arraybackslash}p{#1pt}}
\newcolumntype{y}[1]{>{\raggedright\arraybackslash}p{#1pt}}
\newcolumntype{z}[1]{>{\raggedleft\arraybackslash}p{#1pt}}

\usepackage{pgf}
\usepackage{colortbl}
\usepackage[safe]{tipa} 
\usepackage{tcolorbox}
\usepackage{rotating}
\usepackage[abs]{overpic}
\usepackage{makecell}
\usepackage{longtable}
\usepackage[english]{babel}
\usepackage{csquotes}
\usepackage{hyperref}
\usepackage[most]{tcolorbox}
\definecolor{eventcolor}{RGB}{70,130,180}   
\definecolor{shotcolor}{RGB}{218,165,32}    
\definecolor{entitycolor}{RGB}{220,20,60}   
\definecolor {globalcoloar}{RGB}{150,210,120}   

\usepackage{amsthm, bm, mathtools}
\usepackage{algorithm, algpseudocode}
\usepackage{caption}
\usepackage{placeins} 
\tcbuselibrary{breakable,skins}

\definecolor{gooseyellow}{RGB}{255,236,139}
\tcbset{
  promptboxgoose/.style={
    breakable,
    enhanced jigsaw,
    colback=gooseyellow!40,
    colframe=gooseyellow!70!black,
    boxrule=0.4pt,
    arc=1pt,
    left=6pt, right=6pt, top=4pt, bottom=4pt,
    fonttitle=\bfseries,
    coltitle=black,
    colbacktitle=gooseyellow!80,
    attach boxed title to top left={xshift=6pt,yshift=-\tcboxedtitleheight/2},
    boxed title style={colframe=gooseyellow!70!black,boxrule=0.3pt,arc=1pt,coltitle=black},
    before skip=8pt, after skip=8pt,
    pad at break*=2pt,
  },
}

\newcommand{\modelname}{Aura\xspace}

\newcounter{promptbox}
\renewcommand{\thepromptbox}{\Alph{promptbox}}
\tcbset{
  promptbox/.style={
    enhanced, breakable,
    colback=gray!4, colframe=gray!55!black,
    coltitle=white, colbacktitle=gray!55!black,
    fonttitle=\bfseries\small,
    boxrule=0.4pt, arc=2pt,
    left=6pt,right=6pt,top=4pt,bottom=4pt,
    before skip=8pt,after skip=8pt,
    fontupper=\small\ttfamily\raggedright,
  },
}
\newcommand{\promptlabel}[1]{\refstepcounter{promptbox}\thepromptbox\label{#1}}

\newcommand{\answerTODO}[1][]{\textcolor{red}{\bf [TODO]}}

\usepackage{xspace}

\title{\modelname: Consistent Multi-Subject Video Generation via VLM-Grounded Semantic Alignment}

\author[\dagger]{Zixiang Zhou}
\author{Zhentao Yu}
\author{Yifeng Ma}
\author{Hongmei Wang}
\author{Wenqing Yu}
\author{Cong Wang}
\author{Zilin Yang}
\author{Rui Chen}
\author{Jiarong Ou}
\author{Yezhou Liu}
\author{Yuan Zhou}
\author{Qinglin Lu}

\affiliation{Tencent Hunyuan}

\newcommand{\papertitle}{\modelname: Consistent Multi-Subject Video Generation via VLM-Grounded Semantic Alignment}

\usepackage{fancyhdr} 
\fancypagestyle{titlepagewithlogo}{
  \fancyhf{} 
  \rhead{\includegraphics[height=0.6cm]{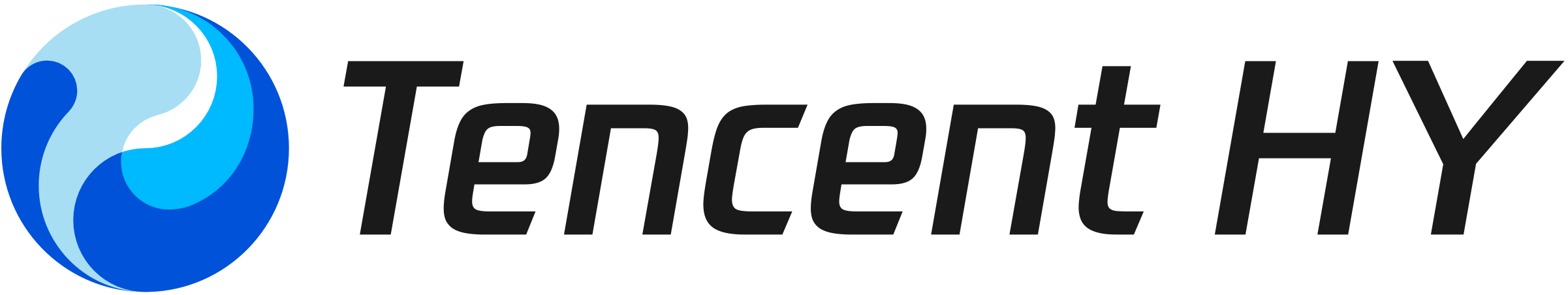}} 
}

\metadata[Project Page]{\url{https://aura-project-page.github.io/}}
\metadata[Code]{\url{https://github.com/Camellia997/Aura}}

\makeatletter
\renewcommand{\mymaketitle}{%
  \tcbset{enhanced,frame hidden}
  \tcbset{left=0.5cm}
  \tcbset{right=0.5cm}
  \tcbset{top=0.5cm}
  \tcbset{bottom=0.5cm}
  \tcbset{arc=10pt}
  \tcbset{colback=white}
  \tcbset{before skip=0pt}
  \tcbset{grow to left by=1.5pt}
  \tcbset{grow to right by=1.5pt}
  \begin{tcolorbox}
    \setlength{\parindent}{0cm}
    \setlength{\parskip}{0.5cm}
    {
      \setlength{\parskip}{0cm}
      \raggedright
      \nohyphens
      {
        \setstretch{1.618}
        \centering\titlelist\par
      }
      \vskip 0.2cm
      \centering\authorlist\par
      \vskip 0.2cm
      \affiliationlist\par
      \contributionlist\par
    }
    \abstractlist\par
    \vskip 0.15cm
    {
      \setlength{\parskip}{0cm}
      \ifdefempty{\metadatalist}{\vspace*{0.65cm}}{\centering\metadatalist\par}
    }
  \end{tcolorbox}
  \tcbset{reset}
  \FloatBarrier
}
\makeatother

\newcommand{\symfootnote}[2]{{\renewcommand{\thefootnote}{#1}\footnotetext{#2}}}

\abstract{} 

\begin{document}
\thispagestyle{titlepagewithlogo}
\maketitle
\symfootnote{\ensuremath{\dagger}}{Corresponding Author: \email{zxzhou0916@163.com}}

\vspace{0mm}
\begin{center}
  \captionsetup{type=figure,hypcap=false}
  \includegraphics[width=\linewidth]{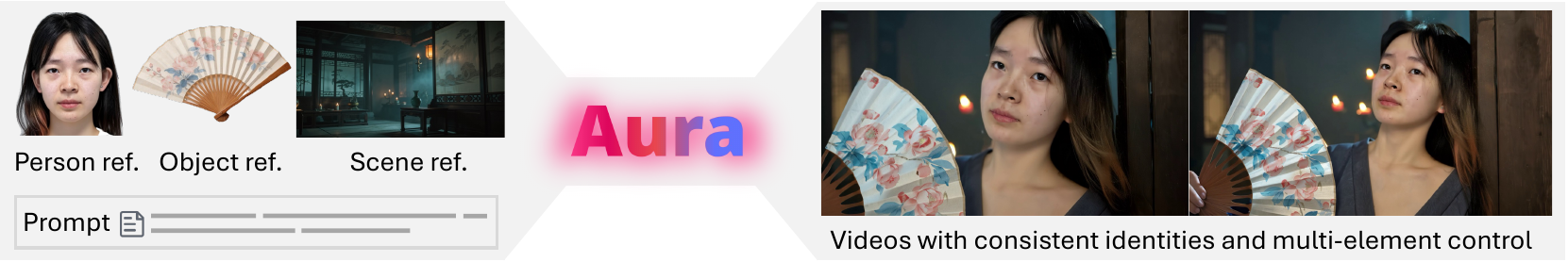}
  \captionof{figure}{\modelname generates high-fidelity videos with strong identity consistency and faithful prompt alignment across a wide range of challenging scenarios, including single-subject animation, multi-subject interaction, and compositional scenes with heterogeneous visual elements (e.g., characters, objects, and backgrounds). \modelname robustly preserves fine-grained subject appearance while producing coherent motion and natural interactions.}
  \label{fig:teaser}
\end{center}

\vspace{2mm}
\begin{center}
{\large\textbf{Abstract}}
\end{center}
  Subject-driven and multi-element video generation are central to controllable video synthesis, but existing methods still struggle to preserve identity consistency and model complex relationships among multiple subjects. In this paper, we propose \modelname, a unified framework for high-fidelity and identity-consistent video generation. To better capture scene dynamics and subject interactions, we introduce AI director-level captions that provide dense and structured descriptions of video content. We further leverage a vision-language model (VLM) with learnable queries to extract multimodal semantic features from textual and visual references, covering both global semantics and fine-grained visual cues. To bridge the representational gap between the VLM and the Diffusion Transformer (DiT), we design a two-stage alignment strategy that progressively maps VLM features into the DiT feature space. For visual conditioning, we adopt token concatenation to inject reference information directly into the generation process. To distinguish heterogeneous subject types and reduce common copy-paste artifacts, we develop a subject-aware RoPE-Shift mechanism. To further differentiate reference images of different categories, we introduce subject-aware learnable tokens. In addition, we introduce Memory Tokens to balance the training signal across examples with different numbers of reference subjects. During inference, Progressive-APG (Adaptive Prompt Guidance) further alleviates oversaturation and improves semantic alignment with user prompts. Finally, we build a high-quality video-subject image dataset through a dedicated data construction pipeline. Extensive experiments show that our method achieves state-of-the-art performance on both single-subject generation and more challenging multi-element scenarios.

\section{Introduction}
\label{sec:intro}

Recent large-scale video diffusion models~\citep{sora,wan2025wan,kong2024hunyuanvideo,hong2022cogvideo} have pushed text-to-video (T2V) synthesis close to real footage on short clips. Yet for most creative applications -- film pre-visualization, digital avatars, e-commerce advertising, and personalized storytelling -- a pure text prompt is no longer a sufficient interface. Users want to specify \emph{who} appears (a particular person, product, or character), \emph{what} the scene contains (reference objects or environments), and \emph{how} the subject speaks or moves (via audio, pose, or other motion cues). This demand has fostered a rapidly growing sub-field we refer to as \emph{subject- and human-centric controllable video generation}, where one or more reference images together with auxiliary modalities jointly steer the generative process on top of a text prompt.

\vspace{2pt}
\noindent\textbf{Entangled bottlenecks.} Despite encouraging progress~\citep{liu2025phantom,li2025bindweave,fei2025skyreels,hu2025hunyuancustom,wang2026refalign,guo2026dreamidomni,zhang2025kaleido}, existing approaches still fall short of delivering truly controllable, high-fidelity, and robust human-centric videos due to several interlocking limitations. Injecting reference features strongly enough to preserve fine-grained identity tends to suppress prompt responsiveness, yielding rigid poses, copied backgrounds, and the widely reported ``copy-paste'' artefact~\citep{liu2025phantom,li2025bindweave,pan2025idcraftervlmgroundedonlinerl}, whereas lighter injection causes visible identity drift. When multiple humans, products, and environments co-occur, current models struggle to bind the right appearance to the right entity, leading to attribute leakage, identity swapping, or omitted subjects~\citep{fei2025skyreels,zhang2025kaleido}. Most pipelines are further trained on clean, studio-like references and degrade sharply on casual inputs with occlusions, motion blur, or non-frontal faces~\citep{guo2026wildactor}. Finally, the standard diffusion / flow-matching denoising loss is a per-token reconstruction objective that offers no explicit supervision on identity similarity or prompt--subject alignment; while DreamID-Omni~\citep{guo2026dreamidomni} and RefAlign~\citep{wang2026refalign} begin to address this gap, a principled modality-aware alignment framework jointly handling identity and text supervision is still missing.

\vspace{4pt}
\noindent\textbf{Our work.} We tackle these issues with \textbf{Aura}, a unified framework for human-centric controllable video generation built on three principles: \emph{balanced multi-modal conditioning} that decouples identity from scene and motion so the prompt retains full expressivity; \emph{compositional multi-reference binding} that routes each reference to its semantic slot across subject counts; and \emph{reference-aware alignment training} that augments denoising with identity-similarity and cross-modal consistency rewards. A purely AIGC \emph{grounding--augmenting--verification} pipeline further supplies high-quality video--reference pairs at scale. Aura consistently surpasses prior arts in identity preservation, prompt following, and motion naturalness, while staying robust to in-the-wild references.

\vspace{4pt}
\noindent\textbf{Contributions.} Our contributions are fourfold: \textbf{1)} a \emph{dual-branch semantic injection} that jointly conditions the DiT on a frozen T5 stream and a Qwen2.5-VL stream via shared-KV cross-attention, with a \emph{T5-teacher alignment} (sentence-level asymmetric InfoNCE plus token-level Hungarian matching) that places VLM meta-queries on T5's manifold and makes parameter-free KV sharing feasible; \textbf{2)} a \emph{subject-aware disambiguation} combining per-category \emph{learnable tokens} with a \emph{Subject-Aware RoPE shift}, giving each reference a feature- and coordinate-level identity to avoid same- and cross-category collisions on the 3D rotary grid; \textbf{3)} an ``Coarse-Align $\rightarrow$ Refine-Align $\rightarrow$ Ref-Only $\rightarrow$ Joint-Mix'' curriculum paired with a \emph{norm-only progressive APG} scheme that stabilizes high-guidance generation along the text and reference axes; and \textbf{4)} a purely AIGC \emph{grounding--augmenting--verification} pipeline that removes the reliance on studio-captured data. Together, these components set a new state of the art on single- and multi-subject benchmarks.

\vspace{-2mm}

\section{Related Work}
\label{sec:related}

Reference-image-conditioned video generation, also known as Subject-to-Video (S2V) or Reference-to-Video (R2V), animates one or more user-provided subjects into a temporally coherent clip that follows a text prompt while preserving each subject's identity~\citep{liu2025phantom,jiang2025vace,li2025bindweave}. Along the \emph{number of reference subjects}, the literature splits into \emph{single-} and \emph{multi-subject} consistency. We exclude audio-driven works, which lie outside our scope.

\vspace{-2mm}
\subsection{Single-Subject Consistency}
\label{subsec:single}

\vspace{-2mm}
Given one reference (most often a portrait), the task is to preserve identity and appearance across the video while obeying the prompt and avoiding \emph{copy--paste} artefacts such as frozen pose or leaked background~\citep{liu2025phantom,pan2025idcraftervlmgroundedonlinerl}. It inherits T2I identity customisation -- from optimisation-based DreamBooth~\citep{ruiz2023dreambooth} and Textual Inversion~\citep{gal2022image} to tuning-free IP-Adapter~\citep{ye2023ip} and PhotoMaker~\citep{li2024photomaker} -- and extends it to video.

The dominant route inserts lightweight adapters or shallow feature concatenations that inject reference tokens into a (partially tuned) video DiT. Face-centric examples include ID-Animator~\citep{he2024id}, ConsisID~\citep{yuan2025identity}, Stand-In~\citep{xue2025standin} and MotionCharacter~\citep{fang2024motioncharacter}, while CustomVideo~\citep{wang2026customvideo} and DisenStudio~\citep{chen2024disenstudio} adopt per-subject tuning. Beyond face-only ID, Phantom~\citep{liu2025phantom} formalises S2V and, on MMDiT, fuses low-level 3D-VAE and high-level CLIP features with a million-scale triplet dataset to suppress the copy--paste shortcut. To remove the \emph{frontal-view bias}, Virtually Being~\citep{xu2025virtually} uses studio multi-view capture, and WildActor~\citep{guo2026wildactor} releases Actor-18M and proposes an \emph{Asymmetric Identity-Preserving Attention} with an \emph{identity-aware 3D RoPE}, reaching SOTA body consistency on Actor-Bench. ID-Crafter~\citep{pan2025idcraftervlmgroundedonlinerl} further uses Qwen2.5-VL to parse prompt and reference jointly and applies VLM-grounded online RL. These methods achieve strong identity fidelity, but shallow concatenation tends to over-copy references and prompt controllability remains reference-sensitive.

\vspace{-2mm}
\subsection{Multi-Subject Consistency}
\label{subsec:multi}

\vspace{-2mm}
With multiple references, the challenge shifts from \emph{``preserve one identity''} to \emph{``bind each identity to its correct role''}, which additionally introduces (i)~\emph{semantic confusion} across subjects~\citep{huang2025conceptmaster,li2025bindweave}, (ii)~\emph{spatial/interaction misassignment}~\citep{li2025bindweave,deng2025magref}, and (iii)~\emph{optimisation conflicts}~\citep{wang2026refalign}.

Phantom~\citep{liu2025phantom} extends to $1$--$4$ subjects via dynamic slot allocation in window attention. VACE~\citep{jiang2025vace} unifies heterogeneous conditions into a Video Condition Unit and becomes a common backbone, while Concat-ID~\citep{zhong2025concat}, Cinema~\citep{deng2025cinema} and ConceptMaster~\citep{huang2025conceptmaster} scale subject counts via attention-based injection and decoupled binding. SkyReels-A2~\citep{fei2025skyreels} generalises ``subjects'' to heterogeneous ``elements'' with A2-Bench, and Kaleido~\citep{zhang2025kaleido} pushes heterogeneity further via a \emph{Reference-RoPE} plus cross-pair synthesis. To move from feature concatenation to \emph{semantic reasoning}, a second family places an MLLM/VLM before the DiT: BindWeave~\citep{li2025bindweave} feeds interleaved prompt--reference tokens through Qwen2.5-VL-7B for subject-aware conditioning, reaching SOTA \textsc{NexusScore} on OpenS2V-Eval; PolyVivid~\citep{hu2025polyvivid}, MAGREF~\citep{deng2025magref} and Cinema~\citep{deng2025cinema} similarly leverage MLLMs for identity embedding, mask-guided parsing, and long-horizon storytelling. Complementarily, RefAlign~\citep{wang2026refalign} observes that the vanilla denoising loss offers no per-subject supervision; drawing on DiT--VFM alignment~\citep{yu2025repa,leng2025repae,wang2025ddt,yoo2025redi,yao2025reconstruction,zhang2025videorepa}, it adds a \emph{reference-alignment loss} that pulls DiT features toward a frozen VFM (DINOv3~\citep{simeoni2025dinov3}\,/\,SigLIP2~\citep{tschannen2025siglip}) and pushes apart different subjects, reaching $60.42\%$ \textsc{TotalScore} on OpenS2V-Eval -- yet its alignment is \emph{static, frame-wise} and vision-only.

\vspace{-2mm}
\subsection{Positioning of Our Work}
\label{subsec:positioning}

\vspace{-2mm}
Prior work evolves from feature injection~\citep{liu2025phantom,jiang2025vace,zhang2025kaleido} and MLLM binding~\citep{li2025bindweave,deng2025magref} to VFM-side alignment~\citep{wang2026refalign}, yet stays shallow, guidance-fragile, and studio-bound. \textbf{\modelname} unifies them via four designs: \emph{dual-branch T5--VLM injection} with a \emph{T5-teacher alignment} (language-side, complementing RefAlign); \emph{per-category learnable tokens} + \emph{Subject-Aware RoPE shift} extending~\citep{guo2026wildactor,zhang2025kaleido} to multi-subject routing; a four-stage curriculum with \emph{norm-only progressive APG}; and a purely AIGC \emph{grounding--augmenting--verification} pipeline.



\vspace{-2mm}
\section{Method}
\label{sec:method}



\subsection{Overview}
\label{sec:overview}

\begin{figure}[t]
    \centering
    \includegraphics[width=\linewidth]{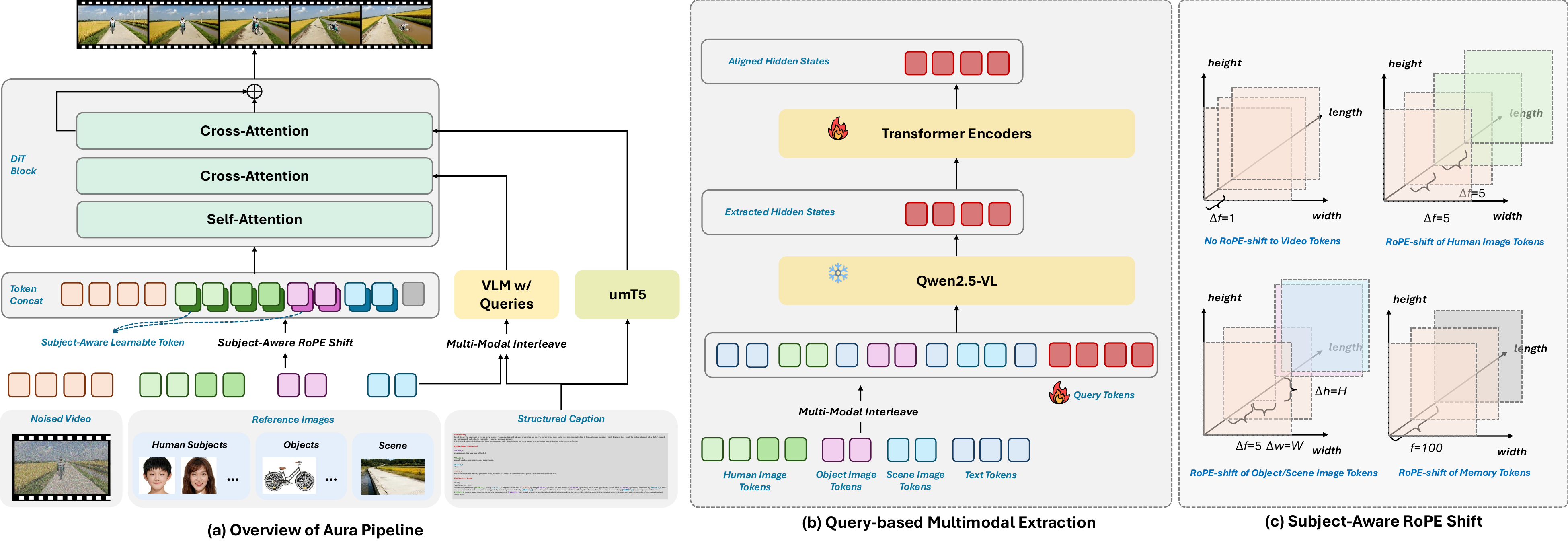}
    \caption{\textbf{Aura architecture.} \textbf{(a)} Noisy video tokens are token-concatenated with references from multiple categories and fed to a DiT with shared-KV cross-attention over T5 and the VLM branch. \textbf{(b)} Learnable meta-queries summarize the interleaved image--text context of a frozen Qwen2.5-VL, projected onto T5's manifold by a zero-init connector. \textbf{(c)} Per-category RoPE shift places categories in mutually disjoint quadrants of the 3D rotary grid.}
    \label{fig:pipeline}
\end{figure}

Aura is a \emph{single} diffusion transformer that unifies T2V and reference-conditioned multi-subject video editing (R2V) with an arbitrary number of references from arbitrary categories (Fig.~\ref{fig:pipeline}(a)). It comprises five coordinated pieces:
\textbf{(1) Reference injection (\S\ref{sec:tokenconcat}):} token-concat with an asymmetric clean-timestep embedding, plus subject-aware disambiguation via per-category learnable tokens and a RoPE shift placing categories in disjoint rotary quadrants.
\textbf{(2) Dual-stream conditioning (\S\ref{sec:vlm}):} query-based multimodal extraction by meta-queries over a frozen Qwen2.5-VL, fused with T5 through a shared-KV cross-attention enabled by T5-teacher alignment.
\textbf{(3) Training (\S\ref{sec:training}):} a four-stage ``Coarse/Fine-Align $\rightarrow$ Ref-Only $\rightarrow$ Joint-Mix'' schedule.
\textbf{(4) Inference (\S\ref{sec:apg}):} a norm-only progressive APG for dual (text/reference) CFG.
\textbf{(5) Data pipeline (\S\ref{sec:datapipeline}):} a subject-centric pipeline mining balanced multi-reference R2V tuples. We describe each of them in detail in the following sections.

\vspace{-2mm}


\vspace{-2mm}
\subsection{Reference Injection}
\label{sec:tokenconcat}

\textbf{Token-concat backbone.} All references are encoded by the shared 3D VAE and mapped to the DiT hidden dim $d$ by an independent 3D patch embedding (spatial patch $2{\times}2$), yielding category-organized tokens $\{z_{\text{ref}}^{(c)}\}_c$ with $c \in \{\text{hum}, \text{obj}, \text{sce}, \text{mem}\}$. We concatenate them with $x_t$ along the sequence axis as $\mathbf{s} = [\, x_t \,\|\, z_{\text{ref}} \,]$ and feed $\mathbf{s}$ to the DiT self-attention, so references influence the output through the \emph{same} full-attention as video tokens, with no extra branch. This raises four liabilities, each pinned to one design choice: (i) \emph{context-length variability} across samples, (ii) \emph{role ambiguity} between noisy video and clean references, and (iii)--(iv) \emph{category and positional ambiguity} across heterogeneous references.

\textbf{(i) Fixed reference budget.} The effective reference count $k$ varies per sample ($1 \le k \le 6$), so $|z_{\text{ref}}|$ would depend jointly on video resolution and $k$. To keep the conditioning budget a function of the video alone, Aura fixes the slot count at $K{=}6$: when $k < K$, the remaining slots are filled by a bank of learnable vectors $\{m_j\}_{j=1}^{K-k}$ tagged with the \emph{memory} category. Both extremes then yield the same $|z_{\text{ref}}|$.

\textbf{(ii) Asymmetric clean-timestep embedding.} Video and reference tokens share the same self-attention yet play different roles---one must be denoised at step $t$, the other is clean context. Since the only per-token knob marking ``which manifold am I on'' is the AdaLN timestep embedding $\mathrm{TE}(\cdot)$, we disambiguate there directly: $\tau_i = \mathrm{TE}(t)$ for $i \in \mathcal{V}$ and $\tau_i = \mathrm{TE}(0)$ for $i \in \mathcal{R}$. A uniform $\mathrm{TE}(t)$ visibly induces color/identity drift on references, which this assignment removes without any tuning.

Beyond (i)--(ii), heterogeneous references on the same concatenated sequence still collide at both the feature and coordinate levels; we disambiguate them along two orthogonal axes.

\textbf{(iii) Subject-aware learnable tokens (feature level).}
\label{sec:learnabletoken}
To let the DiT distinguish ``a human'' from ``a scene'' reference at the \emph{feature} level, we maintain one learnable vector per category (including the memory token):
\begin{equation}
    \mathcal{E} \,=\,
    \{\, e_{\text{hum}},\, e_{\text{obj}},\, e_{\text{sce}},\, e_{\text{mem}} \}
    \subset \mathbb{R}^{d},
    \qquad
    \tilde{z}_{\text{ref}}^{(c)} \,=\, z_{\text{ref}}^{(c)} + e_c.
    \label{eq:cls_add}
\end{equation}
This is the per-token category patch of Fig.~\ref{fig:pipeline}(a): the DiT reads off the category in the first self-attention layer, and $e_{\text{mem}}$ is shared by the memory slots introduced above.

\textbf{(iv) Subject-aware RoPE shift (coordinate level).}
\label{sec:rope}
Eq.~\eqref{eq:cls_add} tags references at the feature level, but concatenation still places all tokens on the \emph{same} 3D rotary grid. Aura resolves the positional collision with a hard-coded per-category offset on the rotary axes, so categories occupy \textbf{mutually disjoint quadrants} (Fig.~\ref{fig:pipeline}(c)). Indexing tokens by $(t, h, w)$ and letting 3D RoPE~\citep{su2024roformer} apply $R_t, R_h, R_w$ channel-wise, for $c \in \{\text{vid}, \text{hum}, \text{obj}, \text{sce}, \text{mem}\}$ we assign a constant shift $\Delta^{(c)} = (\Delta t^{(c)}, \Delta h^{(c)}, \Delta w^{(c)})$:
\begin{equation}
    \mathrm{RoPE}_{3D}^{\text{shift}}\!\big(q;\, t, h, w,\, c\big)
    \;=\;
    \mathrm{RoPE}_{3D}\!\big(q;\,
        t + \Delta t^{(c)},\,
        h + \Delta h^{(c)},\,
        w + \Delta w^{(c)}\big).
    \label{eq:shifted_rope}
\end{equation}
Geometrically: same-category references are staggered only along $t$ (instances remain distinguishable), different categories are separated by full-picture spatial shifts (human/object/scene never collide), and memory slots are pushed to a far temporal position, fully decoupled from real content. Three properties follow. \textbf{(a) Positional non-collision.} Since 3D RoPE is sensitive only to \emph{relative} position, the quadrant layout makes every cross-category offset non-zero and distinct. \textbf{(b) Category-, not instance-, level.} The spatial offset depends only on $c$; within a category only $t$ is incremented per instance, so disambiguation within a category is delegated to appearance, enabling generalization to unseen subject counts. \textbf{(c) Graceful degradation.} With no reference (T2V), Eq.~\eqref{eq:shifted_rope} reduces to standard 3D RoPE, so Aura's T2V pathway is numerically identical to the pretrained baseline. Together with (iii), this yields a \textbf{feature + coordinate dual identity signal}; our ablations (\S4) show removing either component leaves residual identity cross-talk.

\subsection{Dual-Stream Conditioning}
\label{sec:vlm}

On top of $\mathbf{s}$, Aura builds a \emph{separate multimodal semantic pathway} that jointly encodes references and the text prompt into semantic vectors $e_{\text{vlm}}$ consumed by the DiT cross-attention (Fig.~\ref{fig:pipeline}(b)). Neither encoder alone suffices: T5-only leaves no channel for prompt--reference binding and drifts identities under multi-subject prompts, while VLM-only perturbs the pretrained distribution and degrades prompt following and motion. Keeping both decouples two roles---T5 as a \emph{zero-drift anchor}, VLM as a \emph{multimodal broadener}---fused through three coordinated designs: (i) query-based extraction, (ii) T5-teacher alignment, and (iii) a shared-KV cross-attention the alignment makes feasible.

\textbf{(i) Query-based multimodal extraction.}
\label{sec:vlm_core}
Human, object, and scene references are interleaved with the text prompt in a fixed order, with learnable meta-queries $\mathcal{Q} = \{q_k\}_{k=1}^{N_q}$ appended:
\begin{equation}
    \mathcal{X}_{\text{vl}} \;=\;
    \big[\,\mathcal{I}_{\text{hum}} \,\|\,
           \mathcal{I}_{\text{obj}} \,\|\,
           \mathcal{I}_{\text{sce}} \,\|\,
           \mathcal{T} \,\|\,
           \mathcal{Q} \,\big].
    \label{eq:vl_input}
\end{equation}
$\mathcal{X}_{\text{vl}}$ is processed by a frozen Qwen2.5-VL-3B~\citep{Qwen2.5-VL}; we keep only the meta-query positions of its last-layer hidden states as \emph{Extracted Hidden States} $z \in \mathbb{R}^{N_q \times d'}$. Freezing preserves general image--text understanding, while the meta-queries (a) decouple downstream length from reference number/resolution ($|z|{=}N_q$) and (b) provide a trainable, alignable interface to the opaque VLM output. An 8-layer encoder $\mathrm{Enc}_\phi$ refines $z$ into $\tilde{z} = \mathrm{Enc}_\phi(z)$, and a zero-init MLP projects it to $e_{\text{vlm}} \in \mathbb{R}^{N_q \times d}$, so the VLM branch perturbs the DiT by exactly zero at initialization.

\textbf{(ii) T5-teacher alignment.}
\label{sec:vlm_align}
Before the VLM branch can share cross-attention parameters with T5, $e_{\text{vlm}}$ must lie on the same manifold as $e_{\text{t5}}$. We align at \emph{both} sentence and token levels: InfoNCE alone pools away per-token semantics the cross-attention consumes, while naive token-level distance is ill-posed since T5 and VLM meta-queries have different lengths and positionally misaligned semantics. \emph{Sentence-level asymmetric InfoNCE:} with masked-mean-pooled $\bar{e}$, temperature $\tau$, and T5 under $\mathrm{sg}[\cdot]$, over $B$ pairs
\begin{equation}
    \mathcal{L}_{\text{NCE}} \;=\; -\frac{1}{B}\sum_{i}\log
    \frac{\exp\!\big(\langle \bar{e}^{(i)}_{\text{vlm}},\, \mathrm{sg}[\bar{e}^{(i)}_{\text{t5}}] \rangle / \tau\big)}
         {\sum_{j}\exp\!\big(\langle \bar{e}^{(i)}_{\text{vlm}},\, \mathrm{sg}[\bar{e}^{(j)}_{\text{t5}}] \rangle / \tau\big)},
    \label{eq:infonce}
\end{equation}
gives the global direction match KV sharing requires. \emph{Token-level Hungarian matching:} we solve an assignment $\pi^\star$ on the cosine-distance cost matrix and regress matched pairs,
\begin{equation}
    \mathcal{L}_{\text{Hun}} \;=\; \frac{1}{L_V}\sum_{k=1}^{L_V}
    \Big\| e^{(k)}_{\text{vlm}} - \mathrm{sg}\!\big[e^{(\pi^\star(k))}_{\text{t5}}\big] \Big\|_2^2,
    \label{eq:hungarian}
\end{equation}
yielding an order- and length-invariant correspondence that fits exchangeable meta-queries and anchors per-token centers lost by pooling. The alignment objective combines the two:
\begin{equation}
    \mathcal{L}_{\text{align}} \;=\; \lambda_{\text{NCE}}\, \mathcal{L}_{\text{NCE}} \;+\; \lambda_{\text{Hun}}\, \mathcal{L}_{\text{Hun}},
    \label{eq:align}
\end{equation}
complementary in granularity and orthogonal in gradient geometry; removing either breaks the shared KV projections next.

\textbf{(iii) Shared-KV cross-attention.} With $e_{\text{vlm}}$ and $e_{\text{t5}}$ co-located, we fuse them in the DiT cross-attention through a \emph{single} projection triplet $(W_Q, W_K, W_V)$ applied to both streams:
\begin{equation}
    Q = W_Q h, \quad
    K_s = W_K e_s, \quad
    V_s = W_V e_s, \quad s \in \{\text{t5}, \text{vlm}\},
    \qquad
    h^{\text{out}} = h + \tfrac{1}{2}\mathrm{CA}_{\text{t5}}
                       + \tfrac{1}{2}\mathrm{CA}_{\text{vlm}},
    \label{eq:dualkv_fusion}
\end{equation}
with $\mathrm{CA}_s = \mathrm{softmax}(QK_s^\top/\sqrt{d})V_s$. Sharing \emph{all} KV parameters adds \emph{zero} weights over the pretrained T2V DiT, so the VLM branch inherits---rather than competes with---the T5-conditioned prior, which is precisely what (ii) enables.


\vspace{-2mm}
\subsection{Training Strategy}
\label{sec:training}

\vspace{-2mm}
Aura follows a four-stage ``\textbf{Coarse-Align $\rightarrow$ Fine-Align $\rightarrow$ Ref-Only $\rightarrow$ Joint-Mix}'' schedule, unfreezing at most one module group per stage. \textbf{Stage 1 (Coarse-Align).} The VLM pipeline is pretrained against a frozen T5 teacher on (image, text) pairs with $\mathcal{L}_{\text{align}}$, rapidly dragging the VLM output onto T5's manifold and establishing the prerequisite for Eq.~\eqref{eq:dualkv_fusion}. \textbf{Stage 2 (Fine-Align).} With T5 detached, only the VLM encoder stack (meta-queries, $\mathrm{Enc}_\phi$, connector) is unfrozen under the T2V $\mathcal{L}_{\text{FM}}$, performing fine-grained, generation-aware alignment; the zero-init connector makes the start numerically equivalent to the pretrained T2V DiT. \textbf{Stage 3.1 (Ref-Only).} We re-freeze the VLM stack, fully unfreeze the DiT (including $(W_Q, W_K, W_V)$), and optimize $\mathcal{L}_{\text{FM}}$ on R2V tuples only under Eq.~\eqref{eq:dualkv_fusion}, concentrating capacity on reference-consistent generation. \textbf{Stage 3.2 (Joint-Mix).} Keeping Stage-3.1 freezing, the data mix is switched to T2V$+$R2V (T2V samples leave the reference side empty), restoring prompt-following ability and closing the gap between the two input regimes in a single model.


\vspace{-4mm}
\subsection{Inference: Norm-Only Progressive APG}
\label{sec:apg}

\vspace{-2mm}
Aura exposes \emph{two} independent CFG axes at inference: a \textbf{text-free} axis (text prompt dropped) and a \textbf{reference-free} axis (all references dropped). Naively summing the two deltas with large scales causes over-saturation, color drift, and temporal flicker. We adapt \emph{adaptive projected guidance} (APG)~\citep{sadat2024eliminating} to dual-CFG and simplify it to a \textbf{norm-only} and \textbf{progressive} variant. Let $v_{\emptyset}$, $v_{t}$, $v_{rt}$ denote velocity predictions with no, text-only, and both conditions, and define $\Delta_t = v_t - v_{\emptyset}$ and $\Delta_{rt} = v_{rt} - v_{t}$. For each axis we keep the direction of $\Delta$ and rescale only its norm with per-token clipping:
\begin{equation}
    \tilde{\Delta}_s \;=\;
    w_s\,\frac{\min\big(\|\Delta_s\|,\; \kappa_s\big)}{\|\Delta_s\|}\,\Delta_s,
    \qquad s \in \{t, r\},
    \label{eq:norm_only_apg}
\end{equation}
with schedule-dependent norm cap $\kappa_s$, and compose via standard dual-CFG as $v^\star = v_{\emptyset} + \tilde{\Delta}_t + \tilde{\Delta}_{rt}$. Compared with full APG, our variant drops the parallel/orthogonal decomposition (numerically sensitive in the video regime), retains the stabilizing effect of bounded guidance magnitude, and needs only one extra hyper-parameter per axis. Eq.~\eqref{eq:norm_only_apg} yields visibly cleaner long videos at high guidance scales and is used for all reported results.

\vspace{-4mm}
\subsection{Dataset Pipeline}
\label{sec:datapipeline}

\begin{figure}[t]
    \centering
    \includegraphics[width=\linewidth]{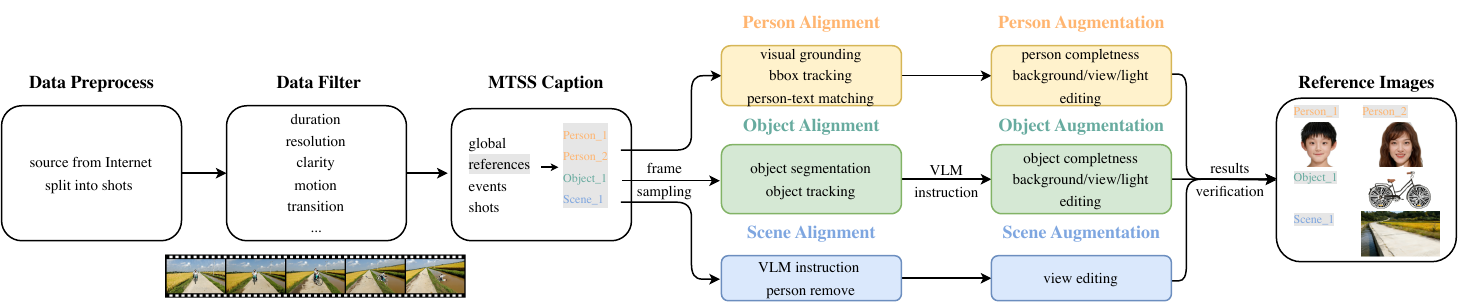}
    \caption{\textbf{Data pipeline.} Starting from raw long videos, our pipeline performs shot segmentation, quality filtering, and AI director-style re-captioning, then constructs three reference streams---human, object, and scene---to produce curated training tuples of \emph{(clip, caption, reference set)}.}
    \label{fig:data_pipeline}
\end{figure}

\vspace{-2mm}
Our curation pipeline turns raw long videos into training tuples of \emph{(clip, caption, reference set)}, as illustrated in Figure~\ref{fig:data_pipeline}. Raw videos are first cut into single-shot clips of $3$--$6$ seconds and filtered by visual sharpness, aesthetic score~\cite{schuhmann2022aesthetic} and optical-flow magnitude, and each surviving clip is re-captioned by an AI director-style captioner~\cite{team2026script} that jointly describes subjects, objects, scene layout and camera behavior, providing grounded phrases for downstream reference construction.

For each clip we construct three complementary reference streams, all sharing a common schema. \emph{(i) Human references:} persons are detected with YOLO~\cite{redmon2016yolo} (occluded faces rejected) and aligned to caption phrases via BLIP-2~\cite{li2023blip2}; crops are edited to diversify background, lighting and clothing, and filtered by ArcFace~\cite{deng2019arcface} similarity to preserve identity. \emph{(ii) Object references:} caption-guided object crops are edited to vary background, lighting, pose and complete truncated silhouettes, with BLIP-2 image--image similarity rejecting drifted edits. \emph{(iii) Scene references:} we prompt HunyuanImage~3.0~\cite{tencent2025hunyuanimage3} to remove foreground subjects and reconstruct clean scene images, then re-render them under varied virtual camera parameters to yield multi-viewpoint scene references. 

\section{Experiments}
\label{sec:experiments}

\vspace{-2mm}
\subsection{Dataset}

\vspace{-2mm}
\textbf{Training set.} Using the pipeline in \S\ref{sec:datapipeline}, we build $\sim$15M clip-level tuples from three complementary sources---cinematic films/TV series, short-form dramas, and user-generated short videos---covering diverse styles, shot compositions, and motion patterns. All clips are $\geq$720P, with aspect ratios spanning $16{:}9$ to $9{:}16$ to enable joint training across landscape, portrait, and intermediate layouts.

\textbf{Test set.} We construct a benchmark of $50$ hand-crafted cases covering multiple orthogonal axes, including \emph{scene context}, \emph{shot scale}, \emph{camera movement}, \emph{visual style}, and \emph{subject category}, each providing at least two categories of reference images (e.g., human$+$object) to probe multi-subject consistency and compositional controllability alongside single-subject fidelity.


\vspace{-2mm}
\subsection{Implementation Details}

We build Aura on the Wan2.2 DiT backbone and train on GPUs with a global batch size of $288$ clips, using AdamW at a constant learning rate of $1\!\times\!10^{-6}$ after a $500$-step linear warmup. To fit the DiT, VLM extractor, and frozen T5 teacher in memory, we shard parameters and gradients via \emph{fully-sharded data parallel} (FSDP). To avoid NCCL stragglers from shape/length skew, we further adopt \emph{three-axis bucketing}---by \emph{aspect ratio} ($16{:}9$ to $9{:}16$), \emph{reference count}, and \emph{reference-category composition} (human/object/scene mix)---drawing each global batch from a single bucket so that all ranks process identically-shaped tensors with comparable compute, stabilizing throughput and gradient synchronization.


\vspace{-2mm}
\subsection{Evaluation Metrics}

\vspace{-2mm}
We evaluate all methods on OpenS2V-Eval~\cite{yuan2025opens2v} using seven metrics normalized to $[0,100]$: \emph{AES}~\cite{schuhmann2022aesthetic} rates per-frame aesthetic appeal; \emph{Motion Smoothness} measures temporal stability via Q-Align~\cite{wu2023dover}; \emph{Motion Amplitude} quantifies motion through optical flow magnitude; \emph{FaceSim-Cur}~\cite{huang2020curricularface} checks identity preservation (with Hungarian matching for multi-subject cases); \emph{GmeScore}~\cite{zhang2024gme} scores text--video faithfulness in a unified multimodal embedding space; \emph{NexusScore} verifies fine-grained subject consistency by comparing YOLO-World~\cite{Cheng2024YOLOWorld} crops against references via GME similarity; and \emph{NaturalScore} judges physical common sense with a GPT-4o~\cite{achiam2023gpt} 1--5 rating. The \emph{Total Score} uses the official open-domain weighting ($0.16, 0.06, 0.02, 0.20, 0.12, 0.20, 0.24$ in the above order), emphasizing subject consistency, identity and naturalness.

We further complement these with a VLM-based evaluation along four dimensions---\emph{action completeness}, \emph{subject consistency}, \emph{video style}, and \emph{camera movement}---using \emph{Gemma4-31B} to rate each on a discrete scale with a short justification.

\vspace{-2mm}
\subsection{Results}

\begin{table}[t]
    \centering
    \caption{Quantitative comparison. \textbf{Left (OpenS2V-Eval)}~\cite{yuan2025opens2v}: metrics in $[0,100]$, \emph{Total} is the official weighted score. \textbf{Right (VLM-based):} \emph{Gemma4-31B} $1$--$5$ ratings. Higher is better; \textbf{bold}/\underline{underline} mark the best/second-best per column.}
    \label{tab:main_eval}
    \renewcommand{\arraystretch}{1.1}
    \setlength{\tabcolsep}{3pt}
    \resizebox{\linewidth}{!}{%
    \begin{tabular}{l cccccccc | cccccc}
        \toprule
        & \multicolumn{8}{c|}{OpenS2V-Eval} & \multicolumn{6}{c}{VLM-based} \\
        \cmidrule(lr){2-9} \cmidrule(lr){10-15}
        Method        & Total             & AES               & MotionSmooth      & MotionAmp         & FaceSim-Cur       & GmeScore          & NexusScore        & NaturalScore      & Action            & Subject           & Style             & Camera            & ID-Cons           & HardCopy          \\
        \midrule
        Wan2.7           & \underline{59.40} & \underline{61.09} & \textbf{96.23}    & 47.88             & \textbf{59.62}    & \underline{55.93} & 78.98             & 38.60             & \underline{4.25}  & \underline{4.90}  & \textbf{5.00}     & \underline{4.25}  & \textbf{4.911}    & \underline{4.911} \\
        \midrule
        HuMo          & 52.62             & 54.27             & \underline{93.51} & 30.56             & 40.18             & 55.17             & 62.29             & \underline{44.17} & 4.05              & \underline{4.90}  & \textbf{5.00}     & 4.05              & 4.163             & \underline{4.939} \\
        Kaleido       & 49.35             & \underline{61.09} & 86.73             & 31.75             & 24.76             & 52.18             & \underline{82.14} & 25.42             & 3.25              & 4.60              & 4.55              & 2.90              & 3.680             & 4.840             \\
        MAGREF        & 52.11             & 58.92             & 93.18             & \underline{54.02} & 20.50             & \textbf{56.29}    & 79.45             & 38.60             & 3.70              & 4.65              & \textbf{5.00}     & 3.40              & \underline{4.192} & 4.577             \\
        RefAlign      & 50.80             & 52.93             & 91.48             & 25.09             & \underline{45.66} & 55.22             & \textbf{84.48}    & 15.00             & 3.25              & 4.50              & 4.65              & 3.60              & 3.625             & 4.393             \\
        \midrule
        \textbf{Ours} & \textbf{61.01}    & \textbf{61.71}    & 88.21             & \textbf{64.79}    & 38.50             & 53.27             & 71.30             & \textbf{67.50}    & \textbf{4.30}     & \textbf{5.00}     & \underline{4.85}  & \textbf{4.35}     & 3.889             & \textbf{4.981}    \\
        \bottomrule
    \end{tabular}%
    }
\end{table}
\vspace{-2mm}

We compare our method against five state-of-the-art video-generation baselines, namely the T2V backbone \emph{Wan}~\cite{wan2025wan} and four subject-to-video systems \emph{HuMo}~\cite{chen2025humo}, \emph{Kaleido}~\cite{zhang2025kaleido}, \emph{MAGREF}~\cite{deng2025magref} and \emph{RefAlign}~\cite{wang2026refalign}. For a fair comparison, we strictly follow the official implementation of each baseline and align the inference hyper-parameters and input formats (e.g., reference image layout, prompt template, sampling steps and classifier-free guidance scale) with their released configurations, so that all methods are evaluated under their own recommended settings. The generated videos of all methods are then assessed with both OpenS2V-Eval~\cite{yuan2025opens2v} and our VLM-based protocol, and the overall quantitative results are summarized in Table~\ref{tab:main_eval}.

As shown in Table~\ref{tab:main_eval}, Ours attains the best overall performance, clearly leading on the aggregate \emph{Total} over both the T2V backbone and all S2V competitors. Since the official weighting concentrates most of its mass on the four fidelity metrics (\emph{NexusScore}, \emph{FaceSim-Cur}, \emph{NaturalScore}, \emph{GmeScore}), this gain reflects genuine semantic faithfulness rather than low-level tricks. Ours ranks first on \emph{NaturalScore} and \emph{AES}, which we attribute to the three-stream \emph{(human, object, scene)} references and the VLM$+$FLUX.1 Kontext editor that jointly promote physical plausibility and aesthetic quality, and also produces markedly richer motion than the near-static outputs of competing S2V methods. 
The VLM-based protocol corroborates these trends, with Ours topping \emph{Subject Consistency} and \emph{Camera Movement} and staying near the best on the remaining dimensions, cross-validating our subject-fidelity advantage and echoing the director-style captions and multi-viewpoint scene references. 

We also conduct a user study using the \emph{Good – Same – Bad} (GSB) protocol. Please refer to Section~\ref{sec:user_study} for details.

\begin{figure}[t]
    \centering
    \includegraphics[width=\linewidth]{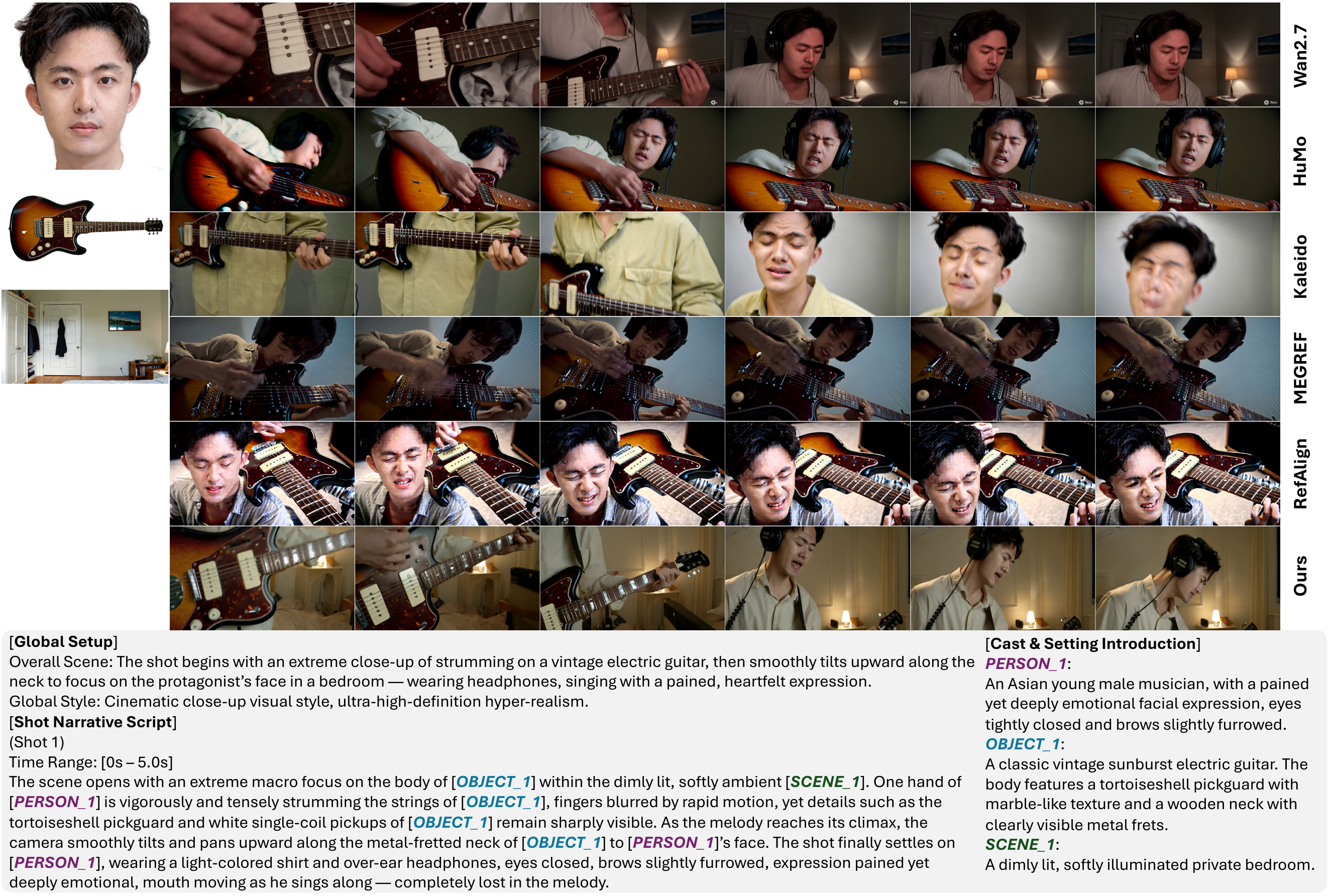}
    \caption{\textbf{Qualitative comparison} with representative multi-subject and multi-element video generation results.}
    \label{fig:results}
\end{figure}

\vspace{-4mm}
\subsection{Ablation Studies}

\vspace{-2mm}
\paragraph{Effectiveness of Dual-Stream Semantic Conditioning}
To verify the effectiveness of our aligned dual-stream semantic conditioning (\S\ref{sec:vlm}), we compare Aura against an \emph{w/o VLM} variant in which the VLM pathway is entirely removed and the DiT is conditioned solely on the T5 embedding of the structured caption, keeping all other components unchanged. The corresponding row appears in the \emph{Dual-Stream Conditioning} block of Table~\ref{tab:ablation_combined} (a). As shown there, removing the VLM pathway mainly hurts aesthetics, motion quality and overall naturalness, while the purely text-driven variant can still stay competitive on narrow textual-matching metrics; this confirms that the VLM stream contributes semantic grounding and visual-world priors that pure T5 conditioning cannot supply.

\vspace{-4mm}
\paragraph{Effectiveness of Training Strategy}
To verify the effectiveness of our four-stage training schedule (\S\ref{sec:training}), we ablate two variants that each drop exactly one stage: \emph{w/o Coarse-Align} skips Stage~1 and relies solely on the downstream flow-matching loss to reach the T5 manifold; \emph{w/o Joint-Mix} stops after Ref-Only and is thus trained only on R2V tuples. Results appear in the \emph{Training Strategy} block of Table~\ref{tab:ablation_combined} (b). Skipping Coarse-Align causes a pronounced collapse on identity- and motion-related metrics, indicating that without explicit alignment the VLM embeddings never land on the T5 manifold; dropping Joint-Mix instead degrades naturalness, action completeness and camera controllability, showing that R2V-only training overfits to reference constraints and erodes the T2V priors. Both variants clearly underperform Ours, confirming the complementary roles of Coarse-Align and Joint-Mix.

\vspace{-2mm}
\paragraph{Effectiveness of Inference Strategy}
\label{sec:ablation_inference}
To verify the effectiveness of our \emph{norm-only progressive APG} inference scheme (\S\ref{sec:apg}), we compare three guidance formulations on the \emph{same} trained DiT, varying only the sampler-side update of the guidance delta $\Delta\epsilon = \epsilon_{\text{cond}} - \epsilon_{\text{uncond}}$: \emph{Regular CFG}~\cite{ho2022classifier} uses standard linear extrapolation; \emph{APG}~\cite{sadat2024eliminating} decomposes $\Delta\epsilon$ into parallel/orthogonal components and down-weights the parallel part; \emph{Ours} keeps its direction but clips its norm by an EMA-tracked threshold. The design is motivated by our observation that, under dual-CFG (text$+$reference) at high scales, the dominant failure mode is not saturation but \emph{guidance-norm explosion} in late steps, which orthogonal projection cannot address. As reported in the \emph{Inference Strategy} block of Table~\ref{tab:ablation_combined} (c), Regular CFG lags on motion and naturalness---symptomatic of norm explosion---while orthogonal APG only marginally improves them since it constrains direction rather than magnitude; our norm clipping instead yields the strongest gains on aggregate score, motion amplitude and naturalness, confirming that it directly targets the true failure mode under dual-CFG.

\begin{table}[t]
    \centering
    \caption{Combined ablation on dual-stream conditioning, training and inference strategies. Higher is better; \textbf{bold} marks the per-column best.}
    \label{tab:ablation_combined}
    \renewcommand{\arraystretch}{1.1}
    \setlength{\tabcolsep}{3pt}
    \resizebox{\linewidth}{!}{%
    \begin{tabular}{l cccccccc | cccccc}
        \toprule
        & \multicolumn{8}{c|}{OpenS2V-Eval} & \multicolumn{6}{c}{VLM-based} \\
        \cmidrule(lr){2-9} \cmidrule(lr){10-15}
        Method                    & Total          & AES            & MotionSmooth   & MotionAmp      & FaceSim-Cur    & GmeScore       & NexusScore     & NaturalScore   & Action         & Subject        & Style          & Camera         & ID-Cons        & HardCopy       \\
        \midrule
        \multicolumn{15}{l}{\emph{(a) Dual-Stream Conditioning}} \\
        w/o VLM                   & 58.27          & 50.12          & 75.86          & 55.05          & 32.79          & 53.92          & 92.36          & 54.58          & 4.10           & \textbf{5.00}  & 4.90           & 4.30           & \textbf{4.093} & 4.815          \\
        \midrule
        \multicolumn{15}{l}{\emph{(b) Training Strategy}} \\
        w/o Coarse-Align          & 55.15          & 60.87          & \textbf{92.28} & 28.14          & 2.71           & 50.87          & 83.82          & 66.25          & 3.60           & 4.75           & \textbf{4.95}  & 3.85           & 3.123          & 4.965          \\
        w/o Joint-Mix             & 56.11          & 55.46          & 84.75          & 50.36          & 34.82          & 52.42          & \textbf{92.45} & 39.17          & 3.60           & 4.65           & 4.80           & 4.10           & 3.808          & 4.769          \\
        \midrule
        \multicolumn{15}{l}{\emph{(c) Inference Strategy}} \\
        Regular CFG               & 58.24          & 59.02          & 87.65          & 46.19          & 31.84          & 52.84          & 84.51          & 54.17          & 4.15           & 4.85           & 4.90           & 4.10           & 3.736          & \textbf{4.981} \\
        Regular APG & 59.45          & 54.37          & 85.68          & 50.62          & 32.08          & \textbf{54.64} & 85.61          & 60.42          & \textbf{4.30}  & 4.95           & \textbf{4.95}  & 4.30           & 3.930          & 4.947          \\
        \midrule
        \textbf{Ours}             & \textbf{61.01} & \textbf{61.71} & 88.21          & \textbf{64.79} & \textbf{38.50} & 53.27          & 71.30          & \textbf{67.50} & \textbf{4.30}  & \textbf{5.00}  & 4.85           & \textbf{4.35}  & 3.889          & \textbf{4.981} \\
        \bottomrule
    \end{tabular}%
    }
\end{table}

\vspace{-2mm}
\paragraph{Effectiveness of Objectives Aligning VLM to T5}
To verify the effectiveness of our alignment objective design (\S\ref{sec:vlm_align}), we adopt a \emph{plug-in protocol}: we run only Stage~1 (Coarse-Align) on (image, text) pairs against the frozen T5 teacher, then use the aligned extractor output $e_{\text{vlm}}$ as a drop-in replacement for the T5 embedding of the pretrained Wan2.2 DiT without any DiT finetuning. We compare our full loss $\mathcal{L}_{\text{align}} = \lambda_{\text{NCE}} \mathcal{L}_{\text{NCE}} + \lambda_{\text{Hun}} \mathcal{L}_{\text{Hun}}$ against a \emph{w/o Hungarian} variant that keeps only the sentence-level InfoNCE. As reported in Table \ref{tab:ablation_align}, removing the Hungarian term causes a sharp deterioration across aesthetics, text alignment, naturalness and all VLM-based dimensions, whereas the full loss recovers competitive quality---indicating that sentence-level InfoNCE alone is insufficient and that token-level Hungarian matching is what drives the VLM output onto the T5 token manifold.

\begin{table}[t]
    \centering
    \caption{Ablation on VLM$\to$T5 alignment objectives, using a Stage-1-only plug-in protocol where the aligned extractor directly replaces Wan2.2's T5 encoder for T2V inference. Higher is better.}
    \label{tab:ablation_align}
    \renewcommand{\arraystretch}{1.1}
    \setlength{\tabcolsep}{3pt}
    \resizebox{\linewidth}{!}{%
    \begin{tabular}{l cccccccc | cccccc}
        \toprule
        & \multicolumn{8}{c|}{OpenS2V-Eval} & \multicolumn{6}{c}{VLM-based} \\
        \cmidrule(lr){2-9} \cmidrule(lr){10-15}
        Method                 & Total          & AES            & MotionSmooth   & MotionAmp      & FaceSim-Cur   & GmeScore       & NexusScore     & NaturalScore   & Action         & Subject        & Style          & Camera         & ID-Cons        & HardCopy       \\
        \midrule
        w/o Hungarian          & 39.67          & 42.62          & 94.72          & \textbf{11.96} & 1.79          & 27.95          & \textbf{85.68} & 25.33          & 1.00           & 1.40           & 1.55           & 1.15           & 1.020          & 4.020          \\
        \textbf{w/ Hungarian}  & \textbf{60.99} & \textbf{60.19} & \textbf{98.40} & 3.63           & \textbf{4.97} & \textbf{34.24} & 84.99          & \textbf{97.00} & \textbf{1.05}  & \textbf{1.80}  & \textbf{2.15}  & \textbf{1.25}  & \textbf{1.196} & \textbf{5.000} \\
        \bottomrule
    \end{tabular}%
    }
\end{table}

\vspace{-4mm}

\section{Conclusion}
\label{sec:conclusion}

\vspace{-2mm}
We presented \textbf{Aura}, a unified diffusion transformer for human-centric controllable video generation. Aura couples a \emph{token-concat reference backbone} (asymmetric clean-timestep embedding, per-category learnable tokens, Subject-Aware RoPE shift) with a \emph{dual-stream semantic pathway} fusing a frozen T5 anchor and a Qwen2.5-VL broadener via parameter-free shared-KV cross-attention, enabled by a \emph{T5-teacher alignment} (asymmetric InfoNCE plus Hungarian matching). A four-stage \emph{Coarse\,$\rightarrow$\,Fine\,$\rightarrow$\,Ref-Only\,$\rightarrow$\,Joint-Mix} curriculum, a \emph{norm-only progressive APG} for dual-CFG, and a \emph{grounding--augmenting--verification} pipeline ($\sim$15M tuples) complete the system. Both OpenS2V-Eval and our VLM-based evaluation show that Aura achieves the best overall performance.


\bibliographystyle{abbrvnat}
\bibliography{neurips_2026}


\newpage
\appendix

\section{AI Director-Devel Caption System (\emph{MTSS})}
\label{sec:mtss}

\modelname is trained, conditioned, and evaluated with \emph{director-style} captions we call the \emph{Multi-Stream Scene Script} (MTSS)~\cite{team2026script}, which replaces monolithic prose with a structured, relationally grounded script. Monolithic captions conflate identity, dynamics, and audio in a single linear string, causing identity ambiguity across shots, weak audio--visual alignment, and non-local edits that break the caption's role as a \emph{control interface}. MTSS removes these failure modes via two principles: \textbf{Stream Factorization} into a \emph{Reference} stream (persistent entities), a \emph{Shot} stream (visual segments), an \emph{Event} stream (grounded audio), and a \emph{Global} stream (ambient context); and \textbf{Relational Grounding} via \emph{identity links} (persistent \texttt{ref\_id}s cited across streams, e.g., \texttt{PERSON\_1}, \texttt{OBJECT\_1}, \texttt{SCENE\_1}) and \emph{temporal links} (shared \texttt{time\_range}s plus intra-description timestamps $[t\,\mathrm{s}]$ that pin micro-actions and utterances to one timeline). Concretely, the Reference stream registers each \texttt{person}/\texttt{object}/\texttt{animal}/\texttt{scene} entity with an \texttt{appearance\_anchor} (with clothing/hairstyle/accessory sub-fields for persons); the Shot stream carries a timestamped \texttt{visual\_description}, a \texttt{camera} field, and \texttt{references\_in\_shot}/\texttt{active\_events} links; the Event stream encodes \texttt{dialogue}/\texttt{sfx}/\texttt{music} occurrences tied to a speaker \texttt{ref\_id}; and the Global stream holds \texttt{scene\_description}, \texttt{global\_style}, and \texttt{global\_audio}. A representative example of our MTSS caption is shown in Box~\ref{box:mtss_example}. 

\FloatBarrier

\begin{tcolorbox}[promptboxgoose, title={Box \promptlabel{box:mtss_example}: MTSS Caption Example}]
\small\ttfamily
[Global Setup]

Overall Scene: A chef with the aura of a martial arts master performs exquisite moves in the kitchen, turning the preparation of Xinjiang Big Plate Chicken into a fluid, visually stunning martial arts spectacle.

Global Style: A visual style combining martial arts kung fu cinema with Chinese cuisine, in high-definition and sharp detail. High saturation colors emphasize the glossy red oil and rich sauce of ingredients. The overall tone is dramatic, energetic, and highly cinematic.

[Cast \& Setting Introduction]

PERSON\_1:

A young Asian woman with a martial arts aura. Wearing a traditional Chinese cross-collar chef’s uniform (similar to martial arts training attire), with tight cuffs for ease of movement. Her hairstyle is neat, her eyes sharp and confident.

OBJECT\_1:

A traditional Chinese wide-bladed cleaver. The blade is thick and sharp, with strong metallic texture, gleaming with cold brilliance under light — perfect for showcasing visual tension in 'mid-air grip' and 'lightning-fast cuts'.

SCENE\_1:

A traditional Chinese kitchen. At the center of the frame is an extremely thick, sturdy solid wood cutting board. Behind it, a roaring Chinese high-heat stove emits intense flames. The environment is rich with the atmosphere of bustling kitchen activity, with lighting focused on the cutting board and stove area.

[Shot Narrative Script]

(Shot 1)

Time Range: [0s – 5.0s]

The shot begins with a rapid push-in from close-up; the subtitle <subtitle>Xinjiang Big Plate Chicken</subtitle> appears at the top of the screen and quickly vanishes. [PERSON\_1] in [SCENE\_1] sinks their qi, presses down on the solid wood cutting board with one hand, causing the whole roasted chicken to fly into the air. [PERSON\_1] slices the airborne chicken into pieces with their hands. The chicken pieces land on the board, and [PERSON\_1] assumes a horse stance, using both palms to exert force from a distance, evenly coating the pieces in cooking wine and soy sauce to lock in flavor and color. Finally, [PERSON\_1] kicks sideways to lift [OBJECT\_1], grabs the cleaver handle with lightning speed, and in a flash of blade light, chops scallions and ginger into segments. [PERSON\_1]’s fingers dance as they mix the sauce, then pours the chicken pieces into a wok to stir-fry until red oil emerges.
\end{tcolorbox}

\section{Why Norm-Only Progressive APG}
\label{sec:why_apg}

This section provides the empirical rationale behind the norm-only, progressive, per-axis simplification of APG adopted in \S\ref{sec:apg}. We revisit the two ingredients of the original APG of~\citet{sadat2024eliminating} -- the parallel/orthogonal \emph{direction} decomposition and the \emph{magnitude} clipping -- and ask, under Aura's dual-CFG video setting, whether each is still necessary. Concretely, for every denoising step we log, per latent frame, the full predicted velocities $v_\emptyset, v_t, v_{rt}$ and the two guidance deltas $\Delta_t, \Delta_{rt}$ used in Eq.~\eqref{eq:norm_only_apg}. Statistics below are aggregated over a probe set of $50$ prompts $\times\, 40$ denoising steps $\times\, 13$ latent frames.

\textbf{Observation 1 (direction decomposition degenerates).} For both axes, the guidance delta is almost perfectly orthogonal to the unconditional prediction $v_\emptyset$: as reported in Table~\ref{tab:apg_perp_ratio}, the mean perpendicular-to-total ratio exceeds $0.996$ on both the text and reference axes, and the mean parallel component is more than an order of magnitude smaller than the perpendicular one. The orthogonal projection operator used by APG,
\begin{equation}
    \Delta_s^{\perp}
    \;=\;
    \Delta_s \;-\; \frac{\langle \Delta_s,\, v_{\emptyset}\rangle}{\|v_{\emptyset}\|^2}\, v_{\emptyset},
    \label{eq:perp_proj}
\end{equation}
therefore reduces to the identity up to numerical error; worse, its denominator $\|v_{\emptyset}\|^2$ couples guidance to the (noisy) unconditional norm in high-dimensional video latents, which we observe amplifies per-frame jitter. Ablations in \S\ref{sec:ablation_inference} confirm that \emph{standard APG}, \emph{clip-only}, \emph{keep-parallel}, and \emph{norm-only} are within sampling noise of each other on identity, motion and artifact metrics, i.e., the direction term carries no measurable signal in our regime. This motivates \emph{dropping} the decomposition.

\textbf{Observation 2 (a single norm cap cannot fit the denoising trajectory).} The magnitude of $\|\Delta_s\|$ varies by more than $3\times$ along the schedule. Averaging per-frame norms over the probe set (Table~\ref{tab:apg_norm_schedule}), the reference axis monotonically decays from $67.2\!\pm\!17.4$ at early steps ($t\!\approx\!999{\to}965$) through $31.3\!\pm\!11.0$ at middle steps ($t\!\approx\!965{\to}888$) down to $18.0\!\pm\!6.4$ at late steps ($t\!\approx\!888{\to}492$); the text axis follows the same pattern, $57.5\!\pm\!9.1 \to 21.7\!\pm\!5.8 \to 12.5\!\pm\!3.6$. Taking the APG default static cap $\tau\!=\!27$ as a reference point, this cap acts unevenly across the trajectory: it compresses $60\%$ of the early-step image-axis guidance and $53\%$ of the early-step text-axis guidance (severely attenuating the large-motion signal), is only marginally active in the middle phase ($14\%$ on the image axis, never on the text axis), and is \emph{entirely inactive} over the late phase where $\|\Delta_s\|$ already lies well below $\tau$. Yet it is precisely the late phase that spawns the dirty-face / over-sharpening artifacts we aim to suppress -- and where a static $\tau$ offers no regularization at all. Hence any \emph{time-invariant} norm threshold $\kappa_s\!\equiv\!\kappa$ is structurally misspecified: a small $\kappa$ needed to control late-step artifacts over-clamps early-step guidance and collapses large-scale motion, whereas a large $\kappa$ tuned to preserve early dynamics is a no-op over the second half of the trajectory. Holding every other component fixed, a linearly annealed cap $\kappa_s(t)$ from $50$ to $15$ simultaneously eliminates both failure modes, whereas no constant $\kappa\!\in\![15,50]$ does. This motivates making the norm cap \emph{schedule-dependent}.

\textbf{Observation 3 (the two CFG axes are not exchangeable).} The reference axis is systematically stronger and more heavy-tailed than the text axis: as summarized in Table~\ref{tab:apg_axis_asymmetry}, global norms are $176.0\pm 47.9$ vs.\ $139.8\pm 25.0$ (image is on average $\sim\!47\%$ stronger); the $50$-th percentile of the per-frame norm distribution (aggregated over all prompt$\times$step$\times$frame samples) is $27.8$ vs.\ $20.2$, and the $95$-th percentile -- characterizing the high-magnitude tail -- is $100.0$ vs.\ $83.6$, so the image axis is larger not only in typical value but also in its upper tail; the reference-to-text norm ratio has mean $1.472\pm 0.562$ with range $[0.856, 3.434]$ across prompts. Consequently, the same static cap $\tau\!=\!27$ clamps $51.5\%$ of image-axis steps but only $33.6\%$ of text-axis steps. Imposing a shared $(\kappa, w)$ therefore either under-regularizes the reference axis (leaking reference-side color drift) or over-regularizes the text axis (damping prompt following). This motivates a \emph{per-axis} parametrization, which is also natural given that dual-CFG already treats the two axes as independent guidance channels.

\begin{table}[t]
    \centering
    \small
    \setlength{\tabcolsep}{10pt}
    \caption{\textbf{The guidance delta is near-orthogonal to $v_\emptyset$ on both axes.} Statistics of $\Delta_s$ relative to $v_\emptyset$, aggregated over the probe set ($50$ prompts $\times\, 40$ denoising steps $\times\, 13$ latent frames). The parallel component is more than an order of magnitude smaller than the perpendicular one on both the reference (image) and text axes, so the APG projection operator in Eq.~\eqref{eq:perp_proj} reduces to the identity up to numerical error.}
    \begin{tabular}{lcc}
        \toprule
        \textbf{Metric} & \textbf{Image guidance} & \textbf{Text guidance} \\
        \midrule
        Perpendicular ratio ($\perp$/total) & $\mathbf{0.996 \pm 0.003}$ & $\mathbf{0.998 \pm 0.002}$ \\
        Parallel component (mean)           & $2.9$                      & $2.1$                      \\
        Perpendicular component (mean)      & $37.5$                     & $29.9$                     \\
        \bottomrule
    \end{tabular}
    \label{tab:apg_perp_ratio}
\end{table}

\begin{table}[t]
    \centering
    \small
    \setlength{\tabcolsep}{8pt}
    \caption{\textbf{A static cap $\tau\!=\!27$ clamps the two CFG axes very unevenly across denoising stages.} Per-frame means of $\|\Delta_s\|$ aggregated over the probe set. Early-stage guidance is heavily over-clamped ($60\%/53\%$), while late-stage guidance -- precisely where dirty-face / over-sharpening artifacts arise -- lies well below $\tau$ and is never regularized. No single $\kappa$ fits both ends of the trajectory, motivating the schedule-dependent cap $\kappa_s(t)$ used in Eq.~\eqref{eq:norm_only_apg}.}
    \resizebox{\linewidth}{!}{%
    \begin{tabular}{lcccc}
        \toprule
        & \multicolumn{2}{c}{\textbf{Image guidance}}
        & \multicolumn{2}{c}{\textbf{Text guidance}} \\
        \cmidrule(lr){2-3}\cmidrule(lr){4-5}
        \textbf{Stage}
        & Per-frame mean & Clamped by $\tau\!=\!27$?
        & Per-frame mean & Clamped by $\tau\!=\!27$? \\
        \midrule
        Early  ($t\!\approx\!999{\to}965$) & $\mathbf{67.2 \pm 17.4}$ & \checkmark\ compresses $60\%$ & $\mathbf{57.5 \pm 9.1}$ & \checkmark\ compresses $53\%$ \\
        Middle ($t\!\approx\!965{\to}888$) & $31.3 \pm 11.0$          & \checkmark\ mild $14\%$        & $21.7 \pm 5.8$          & $\times$\ never triggered      \\
        Late   ($t\!\approx\!888{\to}492$) & $18.0 \pm 6.4$           & $\times$\ never triggered      & $12.5 \pm 3.6$          & $\times$\ never triggered      \\
        \bottomrule
    \end{tabular}%
    }
    \label{tab:apg_norm_schedule}
\end{table}

\begin{table}[t]
    \centering
    \small
    \setlength{\tabcolsep}{10pt}
    \caption{\textbf{The two CFG axes have markedly different norm distributions.} The two percentile rows are computed over the empirical distribution of per-frame $\|\Delta_s\|$ across all $50\!\times\!40\!\times\!13$ (prompt, denoising step, latent frame) samples: the $50$-th percentile captures the typical per-frame magnitude, while the $95$-th percentile characterizes the high-magnitude tail. Image-axis guidance is on average $\sim\!47\%$ stronger than text-axis guidance and exhibits a markedly heavier tail (p$95$: $100.0$ vs.\ $83.6$), with large across-prompt variability (ratio range $[0.856, 3.434]$). A shared static cap $\tau\!=\!27$ therefore clamps the two axes at very different rates ($51.5\%$ vs.\ $33.6\%$), motivating a per-axis $(\kappa_s, w_s)$ parametrization.}
    \begin{tabular}{lcc}
        \toprule
        \textbf{Metric} & \textbf{Image guidance} & \textbf{Text guidance} \\
        \midrule
        Global norm (mean)                           & $176.0 \pm 47.9$ & $139.8 \pm 25.0$ \\
        Per-frame norm, $50$-th percentile (median)  & $27.8$           & $20.2$           \\
        Per-frame norm, $95$-th percentile (tail)    & $100.0$          & $83.6$           \\
        Clamp rate at $\tau\!=\!27$                  & $51.5\%$         & $33.6\%$         \\
        Image/Text intensity ratio  & \multicolumn{2}{c}{$\mathbf{1.472 \pm 0.562}$ \; (range $[0.856,\, 3.434]$)} \\
        \bottomrule
    \end{tabular}

    \label{tab:apg_axis_asymmetry}
\end{table}

\textbf{Design.} The three observations pick out a minimal-change simplification of APG: (i) remove the parallel/orthogonal split, since it is a no-op whose sole side effect is numerical sensitivity; (ii) keep magnitude clipping, since it is the only component with measurable effect on artifacts; (iii) let the cap depend on the denoising step and differ across axes, since both the schedule and the axes carry statistically distinct magnitude distributions. This is exactly Eq.~\eqref{eq:norm_only_apg}: for each axis $s \!\in\! \{t, r\}$ we retain the raw direction of $\Delta_s$ and rescale only its norm by $\min(\|\Delta_s\|, \kappa_s) / \|\Delta_s\|$, with a linear schedule on $\kappa_s$ and a single guidance weight $w_s$. The resulting rule has two scalar hyper-parameters per axis (endpoint caps of the linear schedule, or equivalently $\kappa_s$ at $t\!=\!1$ and $t\!=\!0$, together with $w_s$), no projection, and strictly subsumes the constant-cap and symmetric-axis baselines. Under identical sampler, seed and CFG weights, this variant removes dirty-face artifacts in $> \! 90\%$ of probe prompts while leaving motion intensity on VBench-style metrics statistically unchanged, confirming that the removed components of APG were indeed inert and the added components were exactly those demanded by the statistics above.

\section{Impact of Hungarian Matching}
\label{sec:hungarian_matching}

\begin{figure}[t]
    \centering
    \includegraphics[width=0.6\linewidth]{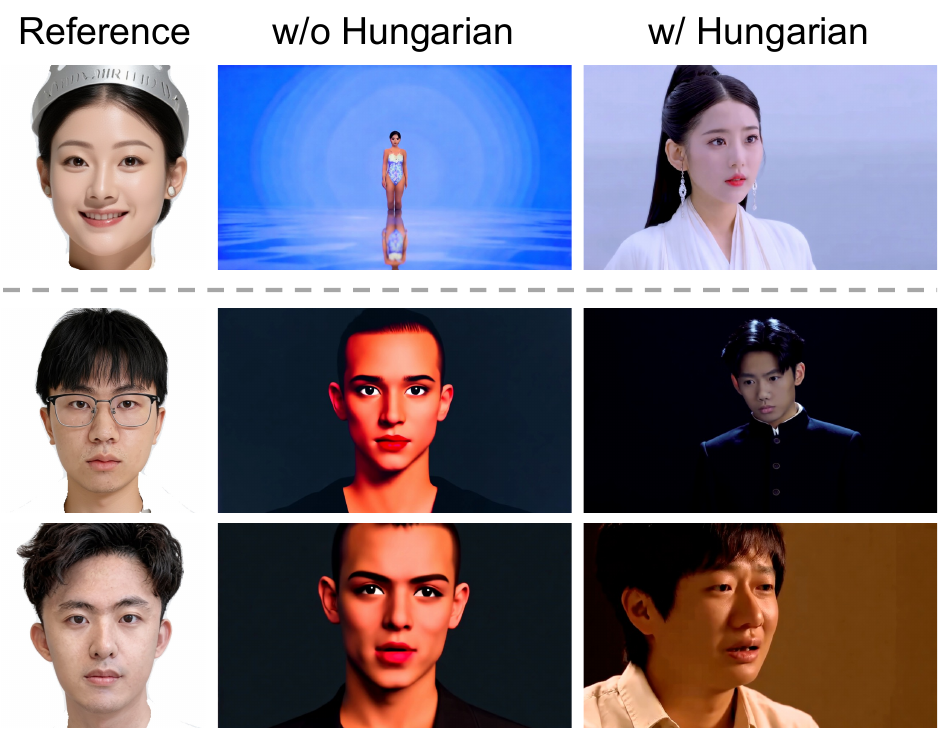}
    \caption{Impact of Hungarian matching on identity alignment. Adding the Hungarian loss improves identity fidelity and preserves more reference-specific facial characteristics under the plug-in protocol.}
    \label{fig:hungarian_loss}
\end{figure}

Figure ~\ref{fig:hungarian_loss} illustrates the impact of incorporating the Hungarian loss into the alignment objective. The comparison follows the same plug-in protocol described in Section 4.5 of the main paper. As shown in the first row, identity injection with the Hungarian loss yields substantially better results than the variant without it. The second and third rows further demonstrate that, without the Hungarian loss, the generated identities tend to converge to similar facial characteristics, whereas incorporating the Hungarian loss enables the model to preserve identity cues that are more faithful to the reference images. These results indicate that the Hungarian loss plays a crucial role in extracting and aligning reference-specific identity information.

\begin{figure}[t]
    \centering
    \includegraphics[width=0.6\linewidth]{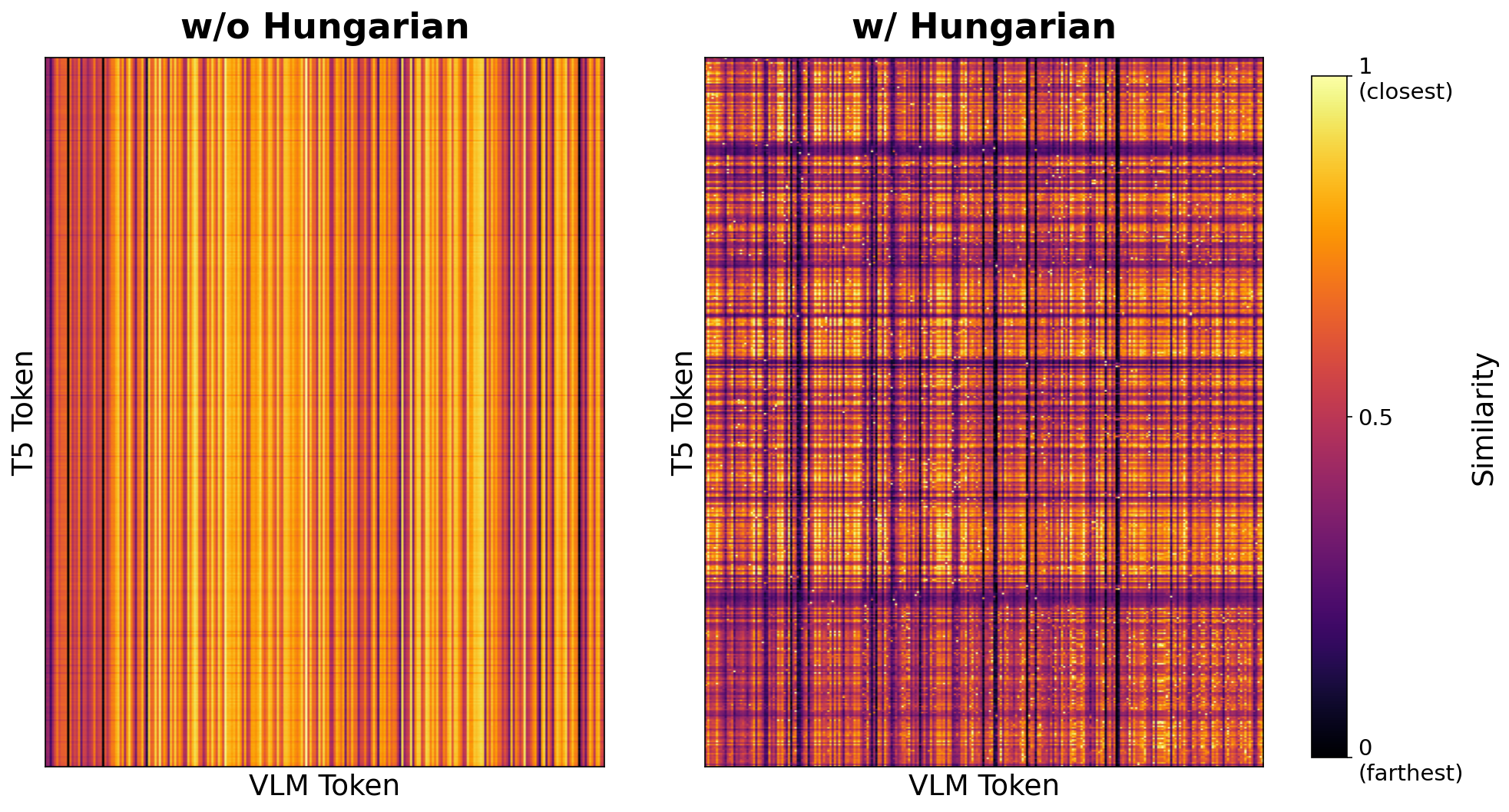}
    \caption{Pair-wise similarity between T5 token and VLM token.}
    \label{fig:grid_figure}
\end{figure}

Figure~\ref{fig:grid_figure} visualizes the token-wise similarity after alignment with Hungarian matching objective. Each matrix visualizes the pairwise similarity between T5 tokens (rows) and VLM tokens (columns), computed as the negated L1 distance and row-normalized to [0, 1] so that the closest VLM token per T5 token appears brightest. Without the Hungarian loss (left), the similarity pattern is dominated by vertical stripes: every T5 token assigns nearly identical similarity scores to VLM tokens, indicating that the feature space is governed purely by VLM-side token properties rather than any meaningful T5–VLM correspondence. With the Hungarian loss (right), a pronounced grid-like structure emerges: both rows and columns develop sharp contrast, meaning that each T5 token selectively attends to a distinct subset of VLM tokens and vice versa. This structured sparsity demonstrates that the Hungarian matching objective successfully drives the model to learn discriminative, token-level alignments between the two modalities, rather than collapsing into a uniform or modality-agnostic representation.

\section{Data Curation Pipeline}
\label{sec:curation_pipeline}

A central difficulty in training Aura is that naive supervision with in-shot reference crops leads to \emph{hard-copy} behavior: the model learns to paste the reference patch verbatim into the output, rather than resynthesizing it under the scene's geometry, lighting, and motion. We therefore design a three-stage curation pipeline that (i) \emph{extracts} an in-shot reference from the source video, (ii) \emph{augments} it via VLM-guided image-to-image editing to break spurious low-level correlations with the target frames, and (iii) \emph{filters} the augmented reference by an identity-preserving similarity check that is tailored to the semantic class (human / object / scene) of the reference. 

\paragraph{Human references.}
For every training video we first run an open-vocabulary visual grounding detector on the caption's person mentions, and track each detected identity across frames with a multi-object tracker, yielding a per-identity tracklet of face/body crops. Given the possibly multiple named entities in the caption, we associate each tracklet with its textual referent by computing the BLIP-2 image--text similarity between each crop and each person-level reference phrase in the caption, and keeping the maximum-similarity match; tracklets whose best similarity falls below a conservative threshold are discarded as unresolved. To prevent the network from collapsing to a copy-paste solution, the selected in-shot crop is further augmented with a VLM-proposed editing plan: a vision--language model is prompted to emit a short, identity-preserving editing instruction along four axes -- background, illumination, clothing, and pose -- which is then executed by an instruction-following I2I model (FLUX.Klein) to synthesize an edited reference. Because such edits can silently drift the facial identity, we perform a strict identity-consistency filter by extracting ArcFace embeddings from the pre- and post-edit crops and retaining only pairs whose cosine similarity exceeds $\tau_{\text{face}}$; all other pairs are rejected. The surviving (edited reference, in-shot video) pairs constitute the human training split.

\paragraph{Object references.}
For object-level references we bypass tracking and directly crop the in-shot object from the video using SAM-3~\cite{carion2025sam3segmentconcepts}, conditioned on the object reference phrase parsed from the caption. Two difficulties arise that are absent for humans. First, the same hard-copy risk motivates a VLM-guided I2I editing step (again via FLUX.Klein) covering background, illumination, and pose perturbations. Second, in-shot object crops are frequently \emph{incomplete}: occlusion by other scene elements truncates the object silhouette, so a naive crop leaks the occlusion pattern as a shortcut feature. We therefore additionally instruct the VLM to propose a \emph{silhouette completion} editing prompt, and use FLUX.Klein's inpainting-style I2I to regenerate the missing contour, producing a fully visible object reference. Since ArcFace is category-specific to faces, identity preservation here is enforced by BLIP-2: we compute the BLIP-2 image embedding of the pre- and post-edit crops and filter out pairs whose cosine similarity is below $\tau_{\text{obj}}$, preventing category drift (e.g.\ a red sedan being edited into a blue SUV) from polluting supervision.

\paragraph{Scene references.}
Scene references require a qualitatively different extraction step, because no in-shot scene crop exists: every frame is partially occluded by foreground subjects. We therefore use HunyuanImage~3.0 to \emph{erase} all foreground humans and objects from a representative frame, producing a clean in-shot background plate. To prevent the network from memorizing the exact viewpoint of this plate, we again issue a VLM-generated editing instruction, but restricted to \emph{camera-level} perturbations -- changes in camera translation and in pan/tilt angles -- which preserve scene identity while forcing the network to reason about view-consistent resynthesis. The edit itself is performed by HunyuanImage~3.0 in a view-conditioned I2I mode. Scene consistency between the pre- and post-edit plates is then verified with a BLIP-2 similarity check analogous to the object case, with threshold $\tau_{\text{scn}}$; failed pairs are discarded.

\paragraph{Summary.}
Across all three categories the pipeline follows the same template -- \textit{extract} $\to$ \textit{VLM-guided I2I editing} $\to$ \textit{semantic-aware consistency filter} -- instantiated with class-appropriate operators. The extraction step guarantees that references are actually \emph{in-shot}, the editing step actively breaks low-level shortcuts (background, lighting, pose, occlusion, viewpoint) that would otherwise invite hard-copy learning, and the filtering step guards each axis of identity (face / object / scene) with the embedding space that is most discriminative for it. 

\begin{figure}[t]
    \centering
    \includegraphics[width=\linewidth]{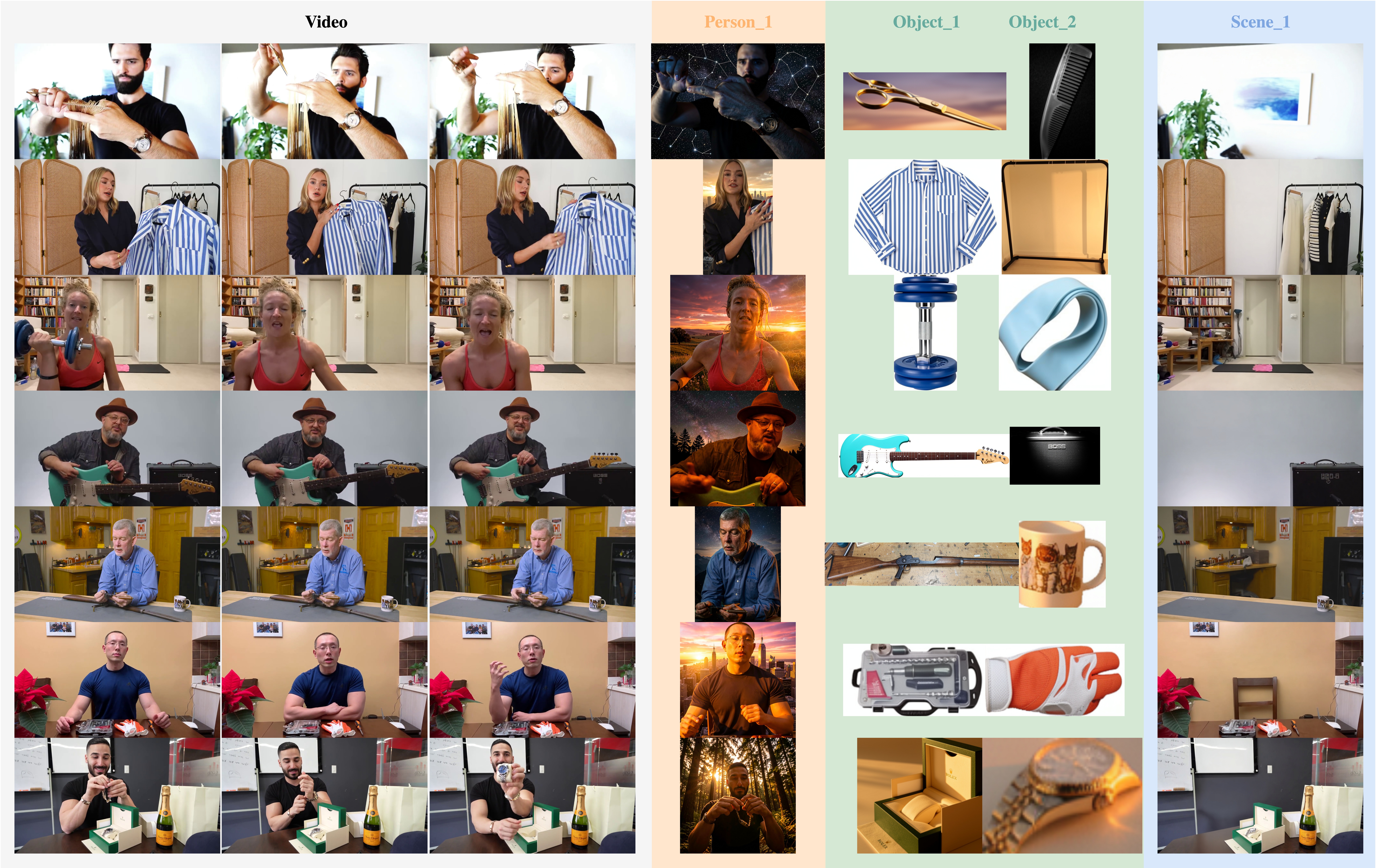}
    \caption{\textbf{Curated reference examples I.} Representative human, object, and scene references produced by our data curation pipeline. The examples show that after the \emph{extract $\to$ edit $\to$ filter} procedure, the curated references preserve the target identity or scene semantics while introducing controlled variations in background, illumination, pose, and viewpoint.}
    \label{fig:data_sample_1}
\end{figure}

\begin{figure}[t]
    \centering
    \includegraphics[width=\linewidth]{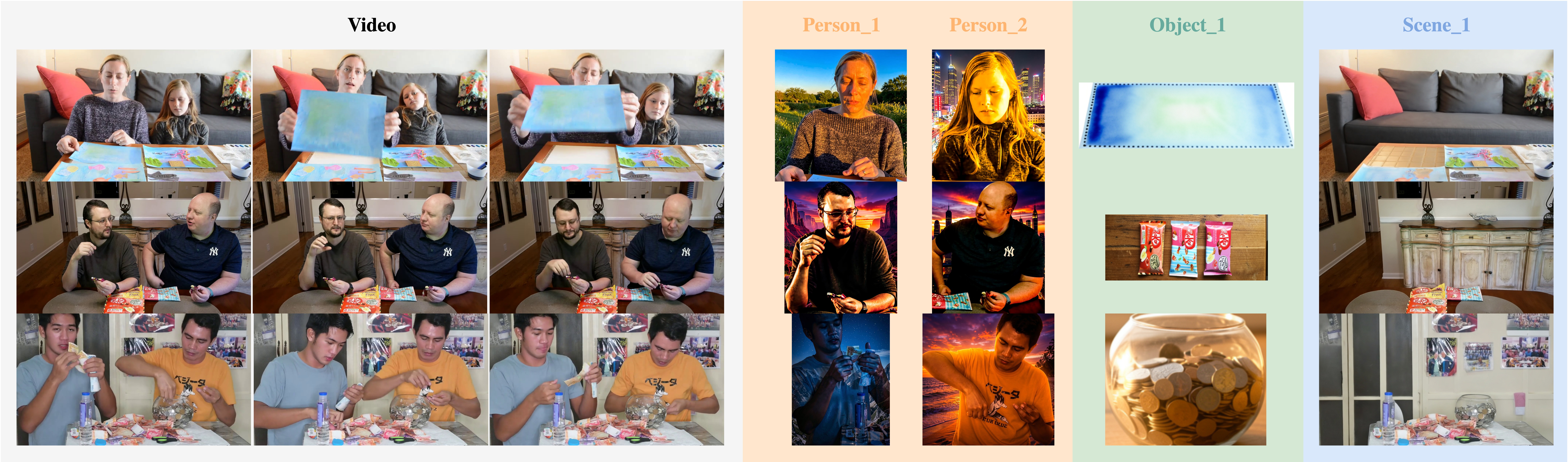}
    \caption{\textbf{Curated reference examples II.} Additional samples from the curated dataset, illustrating diverse human, object, and scene conditions, including occlusion completion, appearance-preserving edits, and view-consistent scene editing. These examples highlight how the pipeline breaks low-level shortcut correlations while retaining semantically faithful supervision for training.}
    \label{fig:data_sample_2}
\end{figure}

To provide a more concrete view of the resulting supervision, Figure~\ref{fig:data_sample_1} and Figure~\ref{fig:data_sample_2} present representative samples produced by our data curation pipeline. These examples illustrate the diversity of the curated human, object, and scene references after the \emph{extract $\to$ edit $\to$ filter} procedure, and show that the final references preserve the target identity or scene semantics while varying low-level factors such as background, illumination, pose, occlusion completion, and viewpoint. This property is exactly what makes the curated data suitable for training Aura to synthesize reference-consistent videos without degenerating into hard-copy behavior.

\section{VLM-based Evaluation}
\label{sec:vlm_eval}

OpenS2V-Eval~\cite{yuan2025opens2v} covers low-level fidelity axes such as aesthetics, face similarity and fine-grained subject retrieval, but is largely blind to several failure modes that matter most for subject-to-video generation: whether the \emph{action} described in the caption is actually carried out, whether the \emph{camera movement} follows the director-style instruction, whether the generated content is stylistically coherent across frames, and whether the model simply \emph{copy-pastes} the reference patch into the output. To complement OpenS2V-Eval we introduce a VLM-based evaluation protocol that uses a frozen vision--language model as an automatic judge and populates the right block of Table~\ref{tab:main_eval} and Table~\ref{tab:ablation_combined}. The protocol is organized into two groups -- \emph{video--text alignment} and \emph{id-consistency} -- that together yield the six columns reported in the main tables.

\paragraph{Judge model and input format.}
We use \emph{Gemma-4-31B} as the VLM judge, run locally with deterministic decoding. For every test case the judge receives: (i) the structured director-style caption used at generation time; (ii) all reference images (human / object / scene) associated with the case; (iii) the generated video, represented by a uniformly sub-sampled set of frames fed to the VLM as image tokens. The $50$ hand-crafted test cases defined in \S\ref{sec:experiments} are identical across methods, so every method is evaluated on the same $50$ (caption, references, video) triples.

\paragraph{Two metric groups.}
The protocol distinguishes a \emph{video--text alignment} group, which is scored \emph{per video} along four axes -- \texttt{action\_completion}, \texttt{subject\_consistency}, \texttt{video\_style}, \texttt{camera\_movement} -- and an \emph{id-consistency} group, which is scored \emph{per referenced entity} along two axes -- \texttt{identity\_consistency} and \texttt{hard\_copy}. Because a single test case may contain multiple entities, id-consistency scores are keyed by entity slot (e.g.\ \texttt{PERSON\_1\_identity\_consistency}, \texttt{OBJECT\_2\_hard\_copy}) and then additionally aggregated by semantic class into \emph{person}, \emph{scene}, and \emph{object} buckets. All six axes use an integer $1$--$5$ scale.

\paragraph{Rating protocol.}
Each axis is scored independently with an axis-specific system prompt that (a) states the rubric, (b) asks the judge to emit an integer score in $\{1,2,3,4,5\}$ together with a one-sentence justification, and (c) forbids cross-referencing other axes. We parse the integer from the judge's response; outputs that violate the format are re-queried, and unresolved cases are marked as missing and excluded from that axis's aggregate. For each (method, axis) pair we retain not only the mean but also the full empirical distribution: count, mean, standard deviation, min, max and the histogram over integer buckets $\{1,\ldots,5\}$. The numbers reported in Table~\ref{tab:main_eval} and Table~\ref{tab:ablation_combined} are the means over the $50$ test cases (or, for id-consistency, over all referenced entities within the $50$ cases).

\paragraph{Axes and rubrics.}
We now specify the rubric used for each axis. Unless stated otherwise, $5$ denotes ``fully satisfied / no observable failure'' and $1$ denotes ``completely violated''.

\textbf{(1) \texttt{action\_completion} (Action).} Does the video execute the action(s) specified by the verb phrases of the caption, in the correct order and without truncation? The rubric penalizes both \emph{missing actions} (the subject never performs the described action) and \emph{truncated actions} (the action begins but is cut off before completion). $5$: all actions fully executed; $4$: all actions present but one is visibly truncated; $3$: the primary action is executed, secondary actions are missing; $2$: only a partial gesture towards the action is visible; $1$: no described action occurs.

\textbf{(2) \texttt{subject\_consistency} (Subject).} Across the generated video, does each referenced subject retain a stable identity in shape, texture and articulation, without morphing, duplication or disappearance? $5$: rock-solid across the full clip; $4$: minor texture drift; $3$: identifiable but with visible shape/texture wobbling; $2$: temporary identity swap or duplication; $1$: the subject is unrecognizable for part of the clip.

\textbf{(3) \texttt{video\_style} (Style).} Does the stylistic look of the video (color palette, lighting, lens feel, post-processing) match the style tag and scene description in the caption, and remain coherent across frames? $5$: stylistically coherent and faithful to the caption; $4$: minor deviation in palette or exposure; $3$: style largely correct but with occasional inconsistent frames; $2$: style drifts noticeably over time; $1$: the output is in a visibly different style than requested.

\textbf{(4) \texttt{camera\_movement} (Camera).} Does the camera trajectory (static, pan, tilt, dolly, orbit, crane, etc.) match the movement clause in the director-style caption? $5$: exact match in direction, pacing and magnitude; $4$: correct type, slightly off in magnitude or pacing; $3$: correct family but with direction confusion (e.g.\ left-vs.-right pan); $2$: a different movement is executed; $1$: the camera is static when motion was requested, or vice versa.

\textbf{(5) \texttt{identity\_consistency} (ID-Cons).} Per referenced entity, does the generated instance match the \emph{identity} depicted in the reference image, rather than merely matching the category label? For \emph{person} entities this covers facial structure, hair, body proportion and distinguishing features jointly; for \emph{object} entities it covers instance-level appearance (specific model, color, material, decoration) as opposed to category (e.g.\ ``a red sedan'' vs.\ the particular sedan in the reference); for \emph{scene} entities it covers the specific locale rather than its generic type. This axis is complementary to FaceSim-Cur: while FaceSim is a pairwise cosine on an ArcFace embedding restricted to faces, \texttt{identity\_consistency} asks the VLM for a holistic judgement that remains meaningful for non-face entities and for conditions where ArcFace is unreliable (profile / occluded / stylized faces). $5$: clearly the same instance; $4$: plausibly the same instance with minor feature drift; $3$: same category but ambiguous instance; $2$: visibly different instance; $1$: the referenced instance is absent.

\textbf{(6) \texttt{hard\_copy} (HardCopy).} Per referenced entity, does the generation \emph{re-synthesize} the reference under the scene's geometry, lighting and motion, rather than pasting the reference image verbatim into the frame? The rubric treats hard-copy behavior as a failure, so a higher score indicates \emph{less} hard-copy, i.e.\ the reference is faithfully rendered under new pose, illumination, viewpoint and motion. $5$: the entity is fully resynthesized with new pose/lighting/motion; $4$: mostly resynthesized with one conspicuously copied region; $3$: several copied regions but the overall frame is new; $2$: most of the entity is a direct paste of the reference patch; $1$: the reference image is inserted essentially unchanged.

\paragraph{Rationale and complementarity to OpenS2V-Eval.}
The six axes are designed to be approximately orthogonal to the seven OpenS2V-Eval metrics. \emph{Action} and \emph{Camera} probe caption-conditional semantics that neither GmeScore (a sentence-level embedding cosine) nor NexusScore (a cropped-subject similarity) is sensitive to. \emph{Subject} and \emph{Style} probe \emph{temporal} coherence at a semantic level, whereas Motion Smoothness only checks low-level pixel stability. \emph{ID-Cons} adds a holistic, per-entity identity check that extends beyond ArcFace-restricted face similarity and is the only id-signal available for \emph{object} and \emph{scene} entities. Finally, \emph{HardCopy} targets a failure mode -- verbatim reference pasting -- that none of the OpenS2V-Eval metrics penalize; on the contrary, hard-copying can \emph{inflate} FaceSim-Cur and NexusScore because the pasted reference trivially maximises similarity. Reporting HardCopy alongside these similarity metrics therefore guards against a pathological optimum in which a model is rewarded for copy-pasting, and is consistent with our data-curation design in \S\ref{sec:curation_pipeline}, which is explicitly built to suppress hard-copy behavior at training time.

\section{VLM Evaluation Prompts}
\label{sec:vlm_eval_prompts}

For reproducibility we include below the exact system prompts used by the VLM judge in \S\ref{sec:vlm_eval}. Box~\ref{box:eval_prompt1} drives the per-entity id-consistency group (axes \texttt{identity\_consistency} and \texttt{hard\_copy}), while Box \ref{box:eval_prompt2} drives the per-video video--text alignment group (axes \texttt{action\_completion}, \texttt{subject\_consistency}, \texttt{video\_style}, \texttt{camera\_movement}).

\section{User Study}
\label{sec:user_study}

\begin{figure}[t]
    \centering
    \includegraphics[width=\linewidth]{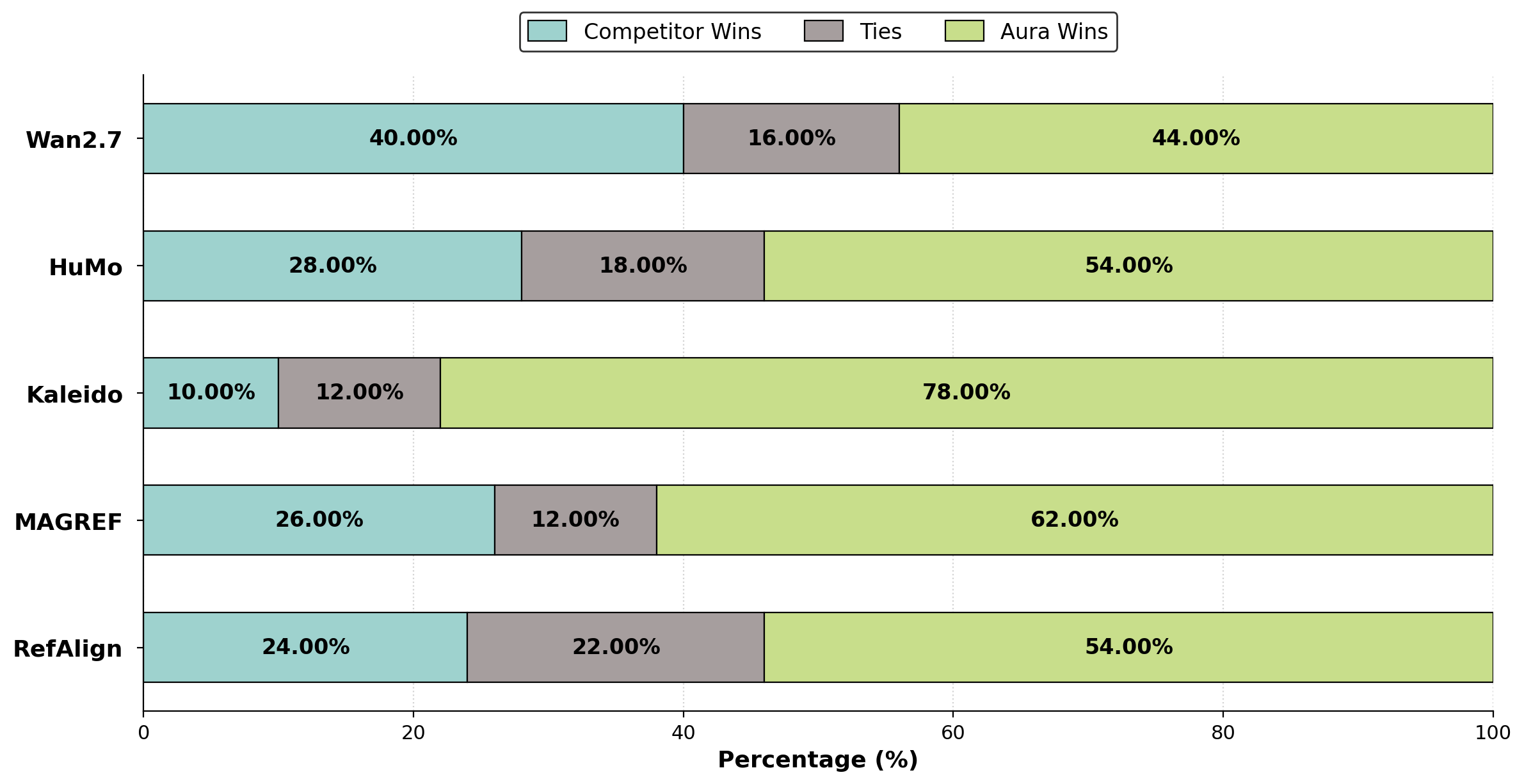}
    \caption{User study results under the GSB protocol. For each competitor we report the percentage of pairs on which Aura wins (\emph{Good}), is judged equivalent (\emph{Same}), or loses (\emph{Bad}). Aura is preferred over every baseline.}
    \label{fig:user_study}
\end{figure}

Beyond the automatic and VLM-based protocols, we conduct a human user study to further compare Aura against five state-of-the-art baselines (\emph{Wan2.7}~\cite{wan2025wan}, \emph{HuMo}~\cite{chen2025humo}, \emph{Kaleido}~\cite{zhang2025kaleido}, \emph{MAGREF}~\cite{deng2025magref}, \emph{RefAlign}~\cite{wang2026refalign}). For each competitor, Aura's video and the competitor's video on the same prompt and references are shown side-by-side in randomized order to blind annotators, who pick the better one along overall quality, subject fidelity, motion plausibility and prompt following under the standard \emph{GSB} protocol -- \textbf{Good} (Aura wins), \textbf{Same} (tie, used whenever the two videos are perceptually indistinguishable), or \textbf{Bad} (competitor wins). Every pair is independently labeled by multiple annotators and consolidated by \emph{majority voting} over \{Good, Same, Bad\} (three-way ties fall back to \textbf{Same}), so each prompt contributes exactly one per-pair GSB outcome, and Figure~\ref{fig:user_study} reports the resulting percentages.

As summarized in Figure~\ref{fig:user_study}, Aura is preferred over \emph{every} baseline, with the \emph{Good} rate consistently exceeding the corresponding \emph{Bad} rate. The advantage is most pronounced against the subject-to-video specialists: Aura wins $78.00\%$ of pairs versus Kaleido (vs.\ $10.00\%$ losses), $62.00\%$ versus MAGREF (vs.\ $26.00\%$), and $54.00\%$ versus RefAlign (vs.\ $24.00\%$), yielding Good$-$Bad margins of $+68$, $+36$ and $+30$ points respectively. In other words, against the S2V specialists annotators prefer Aura by a factor of roughly $7.8\times$ (Kaleido), $2.4\times$ (MAGREF) and $2.3\times$ (RefAlign), indicating that our dual-stream T5--VLM conditioning and director-style MTSS captions translate into clearly perceivable gains in identity preservation, scene richness and camera controllability. Against HuMo, Aura still leads with $54.00\%$ wins against only $28.00\%$ losses (a $+26$-point margin), while the non-trivial tie rate ($18.00\%$) reflects the fact that HuMo already produces reasonable single-subject motion on easier cases. The closest competitor is the T2V backbone \emph{Wan2.7} (Aura $44.00\%$ vs.\ Wan $40.00\%$, ties $16.00\%$): Wan is unconstrained by reference inputs and therefore enjoys maximal visual freedom on open-ended prompts, yet Aura---despite the additional burden of honoring multi-subject references---is still preferred more often, showing that our method injects subject fidelity without sacrificing the generative priors of the backbone. Overall, the human-perceptual verdict is consistent with the quantitative results in Table~\ref{tab:main_eval} and cross-validates Aura's balanced, short-board-free advantage across both T2V and S2V competitors.

\FloatBarrier

\begin{tcolorbox}[promptbox, title={Box \promptlabel{box:eval_prompt1}: Prompt template for the per-video VLM judge}]
\small\ttfamily
You are an expert video-text alignment evaluator. Given a video and a text description, evaluate whether the video content matches the description across the following four dimensions. Score each dimension from 1 (worst) to 5 (best) and provide a reason (no more than 50 words) for each score.

**Evaluation Dimensions:**

1. **Action Completion**: How well the actions/movements described in the text are realized in the video. Consider whether the described actions are fully performed, partially performed, or missing.

2. **Subject Consistency**: Whether the main subjects (people, objects, animals, scenes, etc.) in the video match the text description in terms of appearance, clothing, attributes, and identity.

3. **Video Style**: Whether the visual style, color tone, lighting, mood, atmosphere, and overall aesthetic of the video match the style described in the text.

4. **Camera Movement**: Whether the camera work (tracking, panning, zooming, push-in, pull-out, tilt, bullet-time, etc.) in the video matches what is described in the text.

**Scoring Criteria:**

- 1 = Completely mismatched, no alignment at all
- 2 = Mostly mismatched, only minor elements align
- 3 = Partially matched, some elements align but significant gaps
- 4 = Mostly matched, minor discrepancies only
- 5 = Perfectly matched, all described elements are faithfully realized

**Text Description:**

\{caption\}

Please output your evaluation strictly in the following JSON format and nothing else:

\{\{\\
\hspace*{1em}"action\_completion":~~\{\{\\
\hspace*{2em}"score": <1-5>,\\
\hspace*{2em}"reason": "<no more than 50 words, in English>",\\
\hspace*{1em}\}\},\\
\hspace*{1em}"subject\_consistency":~\{\{\\
\hspace*{2em}"score": <1-5>,\\
\hspace*{2em}"reason": "<no more than 50 words, in English>",\\
\hspace*{1em}\}\},\\
\hspace*{1em}"video\_style":~~~~~~~~~\{\{\\
\hspace*{2em}"score": <1-5>,\\
\hspace*{2em}"reason": "<no more than 50 words, in English>",\\
\hspace*{1em}\}\},\\
\hspace*{1em}"camera\_movement":~~~~~\{\{\\
\hspace*{2em}"score": <1-5>,\\
\hspace*{2em}"reason": "<no more than 50 words, in English>",\\
\hspace*{1em}\}\}\\
\}\}
\end{tcolorbox}

\begin{tcolorbox}[promptbox, title={Box \promptlabel{box:eval_prompt2}: Prompt template for the per-entity VLM judge}]
\small\ttfamily
You are an expert evaluator for assessing identity consistency between reference images and generated video. You are given a video, a text description, and one or more reference images. Each reference image corresponds to a specific subject mentioned in the text (e.g., PERSON\_1, PERSON\_2, OBJECT\_1, SCENE\_1, etc.).\\[4pt]

For EACH reference image provided, evaluate the following two dimensions. Score each from 1 (worst) to 5 (best) and provide a reason (no more than 50 words, in both English and Chinese).\\[4pt]

**Evaluation Dimensions:**\\[4pt]

1. **Identity Consistency**: How well the subject in the video matches the reference image in terms of facial features, body shape, clothing, texture, color, and overall appearance. Note that the text description may intentionally conflict with the reference image (e.g., the reference is a male but the text describes a female character). In such cases, the ideal generation should blend reference identity features (e.g., male facial structure) with the text-described attributes (e.g., female clothing/styling). A high score means the reference identity is recognizable while adapting to the text description.\\[4pt]

2. **Hard Copy**: Whether the subject in the video appears to be an unnatural hard copy/paste of the reference image. If the subject looks overly identical to the reference — to the point of appearing rigid, unnatural, inconsistent with the video context, or lacking proper integration with lighting, pose, and environment — it is considered a hard copy. A HIGH score (5) means NO hard copy issue (natural integration); a LOW score (1) means severe hard copy (looks pasted in).\\[4pt]

**Scoring Criteria:**\\[4pt]
- 1 = Very poor\\[4pt]
- 2 = Poor\\[4pt]
- 3 = Moderate\\[4pt]
- 4 = Good\\[4pt]
- 5 = Excellent\\[4pt]

**Text Description:**\\[4pt]
\{caption\}\\[4pt]

**Reference Images Provided:**\\[4pt]
\{ref\_image\_list\}\\[4pt]

Please output your evaluation strictly in the following JSON format and nothing else:\\[4pt]
\{\{\\
\hspace*{1em}"\{ref\_key\_0\}":~\{\{\\
\hspace*{2em}"identity\_consistency":~\{\{\\
\hspace*{3em}"score": <1-5>,\\
\hspace*{3em}"reason": "<no more than 50 words, in English>",\\
\hspace*{2em}\}\},\\
\hspace*{2em}"hard\_copy":~\{\{\\
\hspace*{3em}"score": <1-5>,\\
\hspace*{3em}"reason": "<no more than 50 words, in English>",\\
\hspace*{2em}\}\}\\
\hspace*{1em}\}\},\\
\hspace*{1em}\ldots\\
\}\}
\end{tcolorbox}

\section{Limitations and broader impact}
\label{sec:limitations}

Despite the encouraging results reported above, Aura has several limitations that we believe point to natural directions for future work.

\paragraph{(1) Residual VLM--T5 misalignment from post-hoc conditioning.}
To inject richer multimodal semantics, Aura augments the backbone with a VLM-based conditioning branch and, to align the VLM stream with the original T5 text stream, introduces a series of alignment and training strategies. These strategies, however, are fundamentally \emph{post-hoc}: the underlying Wan2.2 backbone was pre-trained without ever being conditioned on a VLM stream, so any downstream alignment procedure can only approximate, rather than recover, the joint T5--VLM distribution that a from-scratch pre-training would have learned. As a consequence, on scenes where the VLM-provided semantics disagree subtly with the T5 embedding -- typically long, compositional prompts with fine-grained attribute binding or rare entity names -- we still observe occasional consistency regressions that no amount of post-training completely eliminates. A cleaner solution would be to co-train the VLM and the diffusion backbone from the pre-training stage, which we leave to future work.

\paragraph{(2) The identity--hard-copy trade-off is handled by hand-tuned heuristics.}
A second limitation concerns the fundamental tension between \emph{identity preservation} and \emph{hard-copy suppression} for subject references. Aura mitigates this tension with multiple mechanisms -- the VLM-guided I2I editing step in data curation (\S\ref{sec:curation_pipeline}), the norm-only progressive APG at inference (\S\ref{sec:apg}), and the dual-stream conditioning itself -- and these mechanisms jointly push both axes in the right direction, as confirmed quantitatively (Table~\ref{tab:main_eval}) and perceptually (Figure~\ref{fig:user_study}). However, all of these mechanisms rely on \emph{hard-coded hyper-parameters} (editing strength, filter thresholds $\tau_{\text{face}}/\tau_{\text{obj}}/\tau_{\text{scn}}$, per-axis norm caps $\kappa_s$ and schedules) that are tuned on a held-out probe set and then frozen. In corner cases -- highly stylized portraits, unusual viewpoints, or near-duplicate multi-entity references -- these fixed values can either over-regularize (loss of identity) or under-regularize (lingering copy-paste artifacts), producing failure cases that our current pipeline cannot automatically recover from. A promising direction is to replace these hand-coded knobs with a \emph{learnable} identity--hard-copy controller that adapts per sample, which we expect to improve generalization, especially on out-of-distribution references.

\paragraph{(3) Throughput and distributional drift in the curation pipeline.}
Finally, the data curation pipeline -- although empirically critical for breaking low-level shortcut correlations -- has two practical drawbacks. First, the VLM-guided I2I editing stage (FLUX.Klein / HunyuanImage~3.0) is substantially slower than the rest of the pipeline, and becomes the throughput bottleneck when scaling curation to larger corpora. Second, while the post-edit references satisfy our ArcFace / BLIP-2 consistency filters, a non-trivial fraction of them nonetheless drift outside the distribution of \emph{natural} photographs (e.g., over-smoothed skin, slightly surreal backgrounds, or subtly inconsistent lighting), and we observe that supervising the model with such off-manifold references mildly degrades training stability and downstream fidelity. Addressing this will likely require (i) faster, distillation-based I2I editors, and (ii) an additional ``naturalness'' filter or an adversarial discriminator that rejects post-edit samples whose distribution is too far from real video frames.

\paragraph{Broader impact.}
Aura is obtained by supervised fine-tuning on top of a pre-trained T2V backbone, and therefore \emph{inherits}, rather than introduces, the dual-use risks of that backbone; our fine-tuning neither attempts nor is able to neutralize them. Three risks are most salient. \textit{(i) Deepfakes.} Improved identity preservation, combined with the backbone's high fidelity, could facilitate non-consensual or misleading videos of real individuals. We partially mitigate this at the method level via curation and norm-only APG that discourage hard-copy pasting, and recommend consent-based reference collection, provenance watermarking, and access controls at deployment. \textit{(ii) Biased depictions.} Demographic, cultural, and occupational biases in the backbone persist in Aura's outputs and may even be \emph{amplified} when references themselves encode such priors; SFT should not be interpreted as de-biasing, and users should evaluate Aura on their target demographics before deployment. \textit{(iii) Copyrighted-style imitation.} The backbone can already approximate copyrighted styles or trademarked characters, and subject conditioning can sharpen such imitation when copyrighted material is used as a reference; input-side provenance checks and output-side IP classifiers are necessary complements. On the positive side, the same capabilities enable creative applications (pre-visualization, virtual production, storytelling, education), and we believe the benefits outweigh the risks under the above safeguards.


\FloatBarrier

\end{document}